\documentclass[10pt,twocolumn,letterpaper]{article}

\usepackage{cvpr}              
\definecolor{cvprblue}{rgb}{0.21,0.49,0.74}
\usepackage[pagebackref,breaklinks,colorlinks,allcolors=cvprblue]{hyperref}
\usepackage{siunitx}
\usepackage{booktabs}
\usepackage{tabularx}
\usepackage{array}
\usepackage[accsupp]{axessibility}
\newcolumntype{C}{>{\centering\arraybackslash}X}
\def\paperID{} 
\def\confName{CVPR}
\def\confYear{2026}

\title{REVIVE 3D: Refinement via Encoded Voluminous Inflated prior for Volume Enhancement}

\author{
  \begin{tabular}{@{}c@{}}
    Hankyeol Lee \qquad Wooyeol Baek \qquad Seongdo Kim \qquad Jongyoo Kim\footnotemark[2]
  \end{tabular} \\
  Yonsei University \\
  {\tt\small \{guts4, wooyeol.baek, sdokim07, jy.kim\}@yonsei.ac.kr}
}

\begin{document}

\captionsetup{hypcap=false}
\twocolumn[{
\renewcommand\twocolumn[1][]{#1}
\maketitle
\begin{center}
    \vspace{-10mm}
    \includegraphics[width=\textwidth]{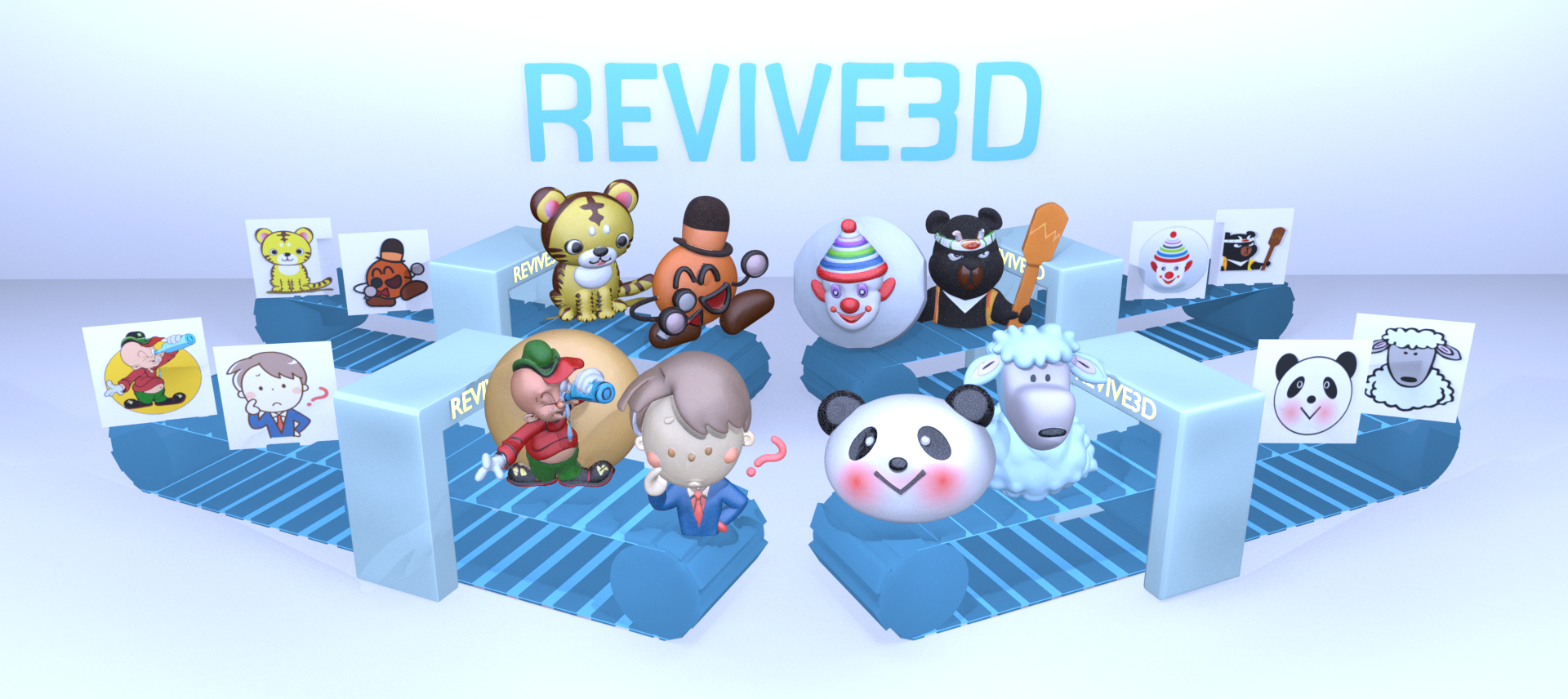}
    \captionof{figure}{REVIVE 3D generates voluminous and detailed 3D meshes from flat images.}
    \label{fig:teaser}
\end{center}
}]
\renewcommand{\thefootnote}{\fnsymbol{footnote}} 
\footnotetext[2]{Corresponding author.} 
\renewcommand{\thefootnote}{\arabic{footnote}}
\begin{abstract}
Recent generative models have shown strong performance in generating diverse 3D assets from 2D images, a fundamental research topic in computer vision and graphics. However, these models still struggle to generate voluminous 3D assets when the input is a \textbf{flat image} that provides limited 3D cues. We introduce \textbf{REVIVE 3D}, a two-stage, plug-and-play pipeline for generating voluminous 3D assets from flat images. In Stage~1, we construct an \textbf{Inflated Prior} by inflating the foreground silhouette to recover global volume and superimposing part-aware details to capture local structure. In Stage~2, 3D Latent Refinement injects Gaussian noise into the Inflated Prior's latent and then denoises it, using the prior's geometric cues to leverage the backbone's pretrained 3D knowledge. Furthermore, our framework supports image-conditioned 3D editing. To quantify volume and surface flatness, we propose \textbf{Compactness} and \textbf{Normal Anisotropy}. We validate Compactness and Normal Anisotropy through a user study, showing that these metrics align with human perception of volume and quality. We show that REVIVE 3D achieves state-of-the-art performance on a challenging flat image dataset, based on extensive qualitative and quantitative evaluations. 
Project page: \url{https://guts4.github.io/REVIVE3D/}
\end{abstract}    
\section{Introduction}

\begin{figure}[t]
  \centering
  \includegraphics[width=1.0\linewidth]{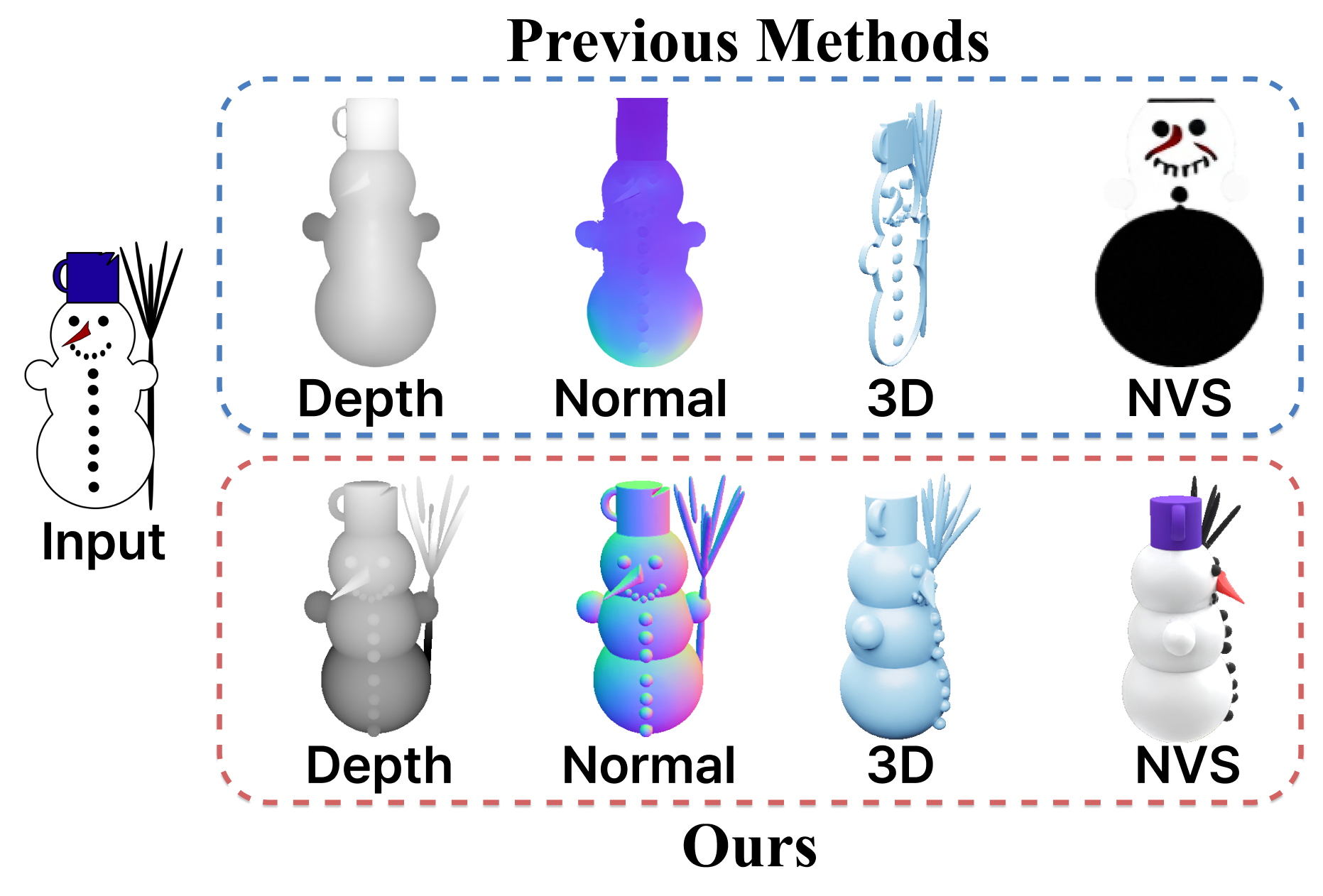}
  \caption{Previous methods fail to generate accurate global structures or details on flat images due to a lack of 3D cues, especially in depth estimation~\cite{yang2024depth}, surface normal estimation~\cite{bae2024dsine}, 3D reconstruction~\cite{hunyuan3d2025hunyuan3d}, and novel view synthesis (NVS)~\cite{liu2023zero}. In contrast, our method successfully recovers detailed 3D geometry.}
  \label{fig:intro_result}
\end{figure}

Recent advances in generating 3D assets from 2D images are crucial for various applications, including virtual reality, video games, and animation. While manual modeling can achieve high-quality 3D assets, it is a time-consuming and labor-intensive process. To address this, numerous high-performance models for automatic single image-to-3D generation have been developed~\cite{wu2024direct3d,hunyuan3d2025hunyuan3d,xiang2024structured}. These methods generate results either by first synthesizing multi-view images as an intermediate step~\cite{liu2023zero, shi2023zero123plus}, or by generating 3D shapes directly from a single image~\cite{hunyuan3d22025tencent, zhao2023michelangelo}.

Despite this progress, even these state-of-the-art single image-to-3D models fail to generate voluminous 3D meshes when the input image lacks 3D cues such as shading, texture gradients, or relative positional information (Fig.~\ref{fig:intro_result}). We define \emph{flat images} as such inputs, including cartoons, line drawings, and flat-shaded art. This scarcity of 3D cues makes it difficult for existing models to estimate depth or surface normals. Furthermore, large-scale training sets for single image-to-3D generation primarily comprise natural photos or rendered views rich in 3D cues. Consequently, flat images are out-of-distribution for these models, leading to failures in both novel view synthesis and direct 3D generation. Prior work has attempted to solve this problem using inflation~\cite{dvorovzvnak2020monster, feng2017magictoon}, 2D-guided pipelines~\cite{zhou2024drawingspinup, peng2024charactergen}, or parametric regression~\cite{luo2023rabit, song2024magiccartoon}. However, inflation-based methods remain silhouette-driven and struggle to recover fine details. Moreover, 2D-guided pipelines face limitations because they supply only 2D guidance, such as depth and pose, which offers no direct 3D guidance. As a result, deformations stay in image space and fail to generate voluminous or back-facing geometry. Parametric approaches are constrained by a model space, yielding restricted results.

To generate voluminous and detailed 3D meshes from flat images, we propose \textbf{REVIVE 3D}, a two-stage, plug-and-play pipeline. Instead of relying on restrictive 2D cues, REVIVE 3D provides direct 3D geometric cues by first constructing a voluminous \emph{Inflated Prior} and then refining this prior in the latent space using a conditional diffusion process. The first stage, \textbf{Inflated Prior Generation}, inflates the foreground silhouette to recover global volume and superimposes part-aware inflations to add local structure, resulting in the Inflated Prior. However, this prior results in \emph{convex-only geometry} because it simply inflates the silhouette and local structure in an additive manner. For example, features that should be concave (like a mouth) or back-facing (like a tail) are rendered only as convex regions. To refine convex-only geometry, our second stage, \textbf{3D Latent Refinement}, leverages a latent diffusion model within a pretrained conditional 3D latent generative backbone~\cite{wu2024direct3d, hunyuan3d2025hunyuan3d} composed of a 3D autoencoder and a diffusion transformer operating in the latent space. In this stage, we first encode the Inflated Prior to obtain its latent, then inject Gaussian noise according to the initial noise level, and denoise it conditioned on the input image. The Inflated Prior provides 3D volumetric cues and part-aware spatial cues about where fine structures should appear. These cues guide the diffusion transformer denoiser within the backbone to recover concavities and refine detailed geometry by leveraging the backbone’s pretrained 3D knowledge.

Through this two-stage pipeline, REVIVE 3D generates voluminous 3D meshes from flat images. 
In our experiments, meshes generated by REVIVE 3D achieve higher Uni3D~\cite{zhou2023uni3d} and ULIP~\cite{xue2023ulip} scores than baselines, indicating strong semantic consistency. However, these metrics fail to quantify the core challenge of generating volume. Therefore, we propose \emph{Compactness} and \emph{Normal Anisotropy} to measure volume and surface flatness. Evaluations with Compactness and Normal Anisotropy show that REVIVE 3D produces more voluminous meshes with lower surface flatness than existing models. These quantitative findings are supported by our user study, which indicates that our method aligns with human perception of volume and quality. Thus, REVIVE 3D overcomes the challenge of generating voluminous 3D meshes from flat images.
\section{Related Work}

\begin{figure*}[t!]
  \centering
  \includegraphics[width=1.0\textwidth]{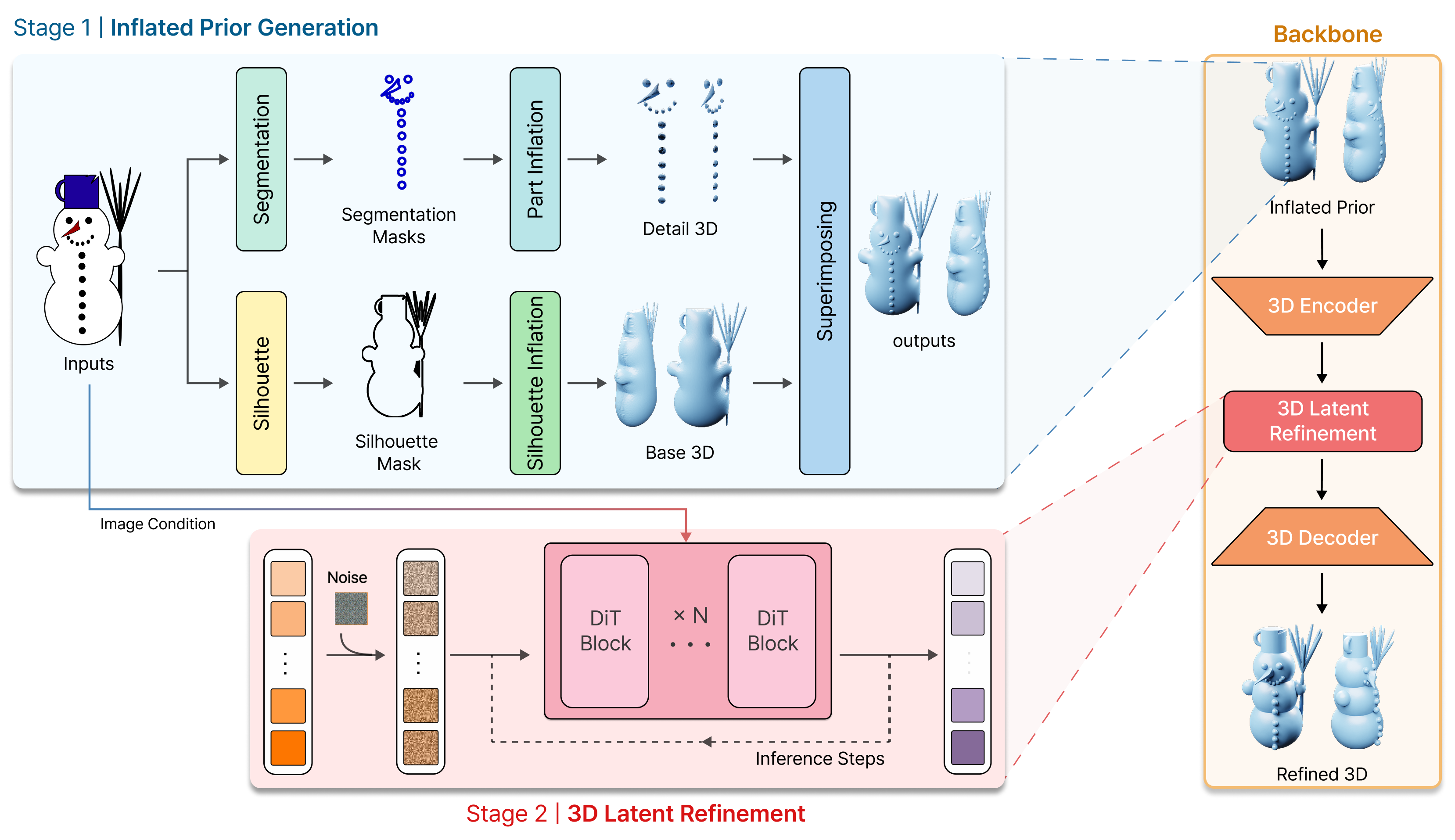}
  \caption{Overview of our method. \textbf{Stage~1} generates the Inflated Prior. We create a Base 3D from the Silhouette Mask and Detail 3D from Segmentation Masks, then combine them via superimposing. \textbf{Stage~2} refines the Inflated Prior by encoding the mesh, injecting noise, denoising it with the image condition, and decoding the result into the Refined 3D mesh.}
  \label{fig:architecture}
\end{figure*}

\paragraph{3D Generative Models.}
3D generative models first leveraged neural implicit representations. NeRF~\cite{mildenhall2021nerf} established differentiable volumetric rendering as the standard backbone. Building on this, score-distillation–based text-to-3D methods optimize 3D content under 2D diffusion priors \cite{poole2022dreamfusion, lin2023magic3d, wang2023prolificdreamer, chen2023fantasia3d, wang2023score}. These pipelines depend on slow iterative optimization and implicit geometry, making mesh extraction difficult. 3D Gaussian Splatting (3DGS)~\cite{kerbl3Dgaussians} introduced explicit 3D Gaussian primitives for fast, differentiable rendering. However, generative models that adopt 3DGS~\cite{chen2024text, yi2024gaussiandreamer, tang2023dreamgaussian, yi2024gaussiandreamerpro} often exhibit geometric inconsistencies under sparse views or text-only conditioning. To improve geometric consistency, image- or text-conditioned multi-view diffusion methods generate view-consistent images that are subsequently lifted to 3D \cite{liu2023zero, shi2023zero123plus, Magic123, sargent2024zeronvs, zhang2025ar, shi2023mvdream, wang2023imagedream, liu2023syncdreamer, sun2023dreamcraft3d}. Subsequent works accelerated multi-view-to-3D reconstruction and output meshes directly \cite{long2024wonder3d, li2023instant3d, xu2024instantmesh, xu2024grm, zheng2024free3d, liu2024one, liu2023one}.

More recently, the field has shifted toward direct 3D generation in compact latent spaces. A 3D autoencoder defines a compact latent space, and a generative model is learned over these latents. At inference, the generative model starts from Gaussian noise in the latent space and iteratively denoises under image conditioning to a clean latent, which the decoder then maps to an explicit 3D representation~\cite{jun2023shap, hong20243dtopia, gupta20233dgen, zhao2023michelangelo, zhang20233dshape2vecset, zhang2024clay, Chen_2025_Dora, ren2024xcube, lan2024ln3diff, wu2024direct3d, hunyuan3d22025tencent, hunyuan3d2025hunyuan3d, xiang2024structured, zeng2022lion}. This approach generates fast, high-quality meshes with improved geometric consistency in recent single image-to-3D pipelines. However, these models still struggle to generate voluminous 3D shapes from flat images.

\paragraph{3D Character Generation.}
To address this limitation, early work on 2D cartoon images relied on foreground silhouette inflation, lifting contours into volumetric meshes~\cite{10.1145/311535.311602, dvorovzvnak2020monster, feng2017magictoon, kraevoy2009modeling}. These methods provide a simple path to apparent volume and animatable geometry. Inflating the foreground silhouette creates front and back surfaces with closed side walls, yielding low-complexity meshes that deform easily. This approach conditions on the visible silhouette, so it offers little guidance for back-facing regions, tends to oversimplify part-level structure, and often relies on manual inputs to resolve ambiguities. To recover missing detail, subsequent research shifted toward 2D-guided pipelines that utilize guidance such as depth or Canny edges, contour removal and refinement, or pose‑conditioned multi‑view cues~\cite{cong2025art3d, zhou2024drawingspinup, peng2024charactergen, chen2023panic, zhang2022creatureshop, han2017deepsketch2face, wang2024nova, he2025stdgen, luo2021simpmodeling}. A parallel line of work performs parametric regression to estimate cartoon-specific pose and shape in a predefined model space~\cite{luo2023rabit, song2024magiccartoon}. These approaches improve local detail fidelity and preserve the character’s identity, yet core limitations remain. 2D-guided pipelines face fundamental limits because the 2D cues are inherently restrictive. For instance, pose guidance constrains the output pose~\cite{peng2024charactergen}, depth guidance relies on unreliable depth from flat images~\cite{cong2025art3d}, and contour methods are specialized for line art~\cite{zhou2024drawingspinup}. Parametric approaches are likewise constrained by a predefined model space, which limits expressiveness for stylized or out-of-template shapes. 

Motivated by these gaps, we work directly in 3D and supply explicit volumetric cues through an Inflated Prior constructed from the input image. The prior captures global volume and part-aware spatial cues specific to the given image, and we refine these latents in 3D without re-projecting to 2D. This design provides direct, image-aligned 3D cues, generating voluminous meshes with detailed structures.
\section{Method}
REVIVE 3D is a two-stage framework that generates voluminous 3D meshes from flat images by first inflating silhouettes and part masks to obtain an Inflated Prior, and then stochastically refining this prior in 3D latent space (Fig.~\ref{fig:architecture}). In Stage~1 (Sec.~\ref{sec:Voluminous}), we inflate the foreground silhouette and part masks to construct the Inflated Prior. Although this prior provides the essential volume, its additive nature leads to convex-only geometry. It is then passed to Stage~2 (Sec.~\ref{sec:refinement}), where we inject Gaussian noise into the 3D latent to correct convex-only geometry. The subsequent denoising process utilizes detailed cues to refine the shape while preserving the volumetric cues established by the prior.

\subsection{Inflated Prior Generation}
\label{sec:Voluminous}
We first recover global shape from the foreground silhouette and then inject local detail from segmented parts, supplying Stage~2 with a volumetric, part-aware prior that guides the refinement of voluminous, detailed 3D meshes.

\paragraph{Global Contour Inflation.}
To generate the Inflated Prior, we adapt the geometric inflation technique from Monster Mash~\cite{dvorovzvnak2020monster}. This technique provides the essential volumetric cues by inflating a 3D volume from the foreground silhouette. Our process begins by extracting the outer contour from the foreground silhouette and then triangulating the 2D planar region enclosed by this contour to form a 2D mesh. We then solve a discrete Poisson equation on this mesh to generate a smooth height field $\tilde h$, which assigns to each vertex a scalar value representing its displacement along the normal direction. For every interior vertex $i$, the equation enforces the local volume constraint:
\begin{equation}
    \sum_{j\in N_i} w_{ij}(\tilde h_j - \tilde h_i) = s_i a_i c, \qquad \tilde h_i = 0 \ \text{for}\ \ i\in\mathcal{C},
    \label{eq:poisson}
\end{equation}
where ${N}_i$ denotes the set of neighboring vertices of $i$, and $w_{ij}$ are cotangent Laplacian weights that capture local curvature between $i$ and its neighbors. The sign term $s_i$ takes $+1$ for front-facing regions and $-1$ for back-facing regions. The lumped vertex area $a_i$ is defined as 1/3 of the total area of triangles incident to vertex $i$, and $c$ controls the global inflation strength. The set of all vertices on the contour $\mathcal{C}$ is fixed to the zero-height plane by the Dirichlet boundary condition.

Finally, to transform the height field produced by the Poisson solve into a smoother surface, we apply a square-root mapping:
\begin{equation}
    h_i = s_i \sqrt{\lvert\tilde h_i\rvert}.
    \label{eq:convex_map}
\end{equation}
The resulting heights $h_i$ represent the displacement of each 2D vertex and define the final 3D vertex coordinates, forming a convex \textbf{Base 3D}.

\paragraph{Local Part Superimposing.}
Silhouette-only inflation often misses fine structures, making it difficult for Stage~2 to produce voluminous, detailed 3D meshes. We therefore add coarse part cues, extracted via an automatic segmentation mask method~\cite{ravi2024sam2}, to provide additional volumetric and local structural guidance for 3D latent refinement. We first filter predicted candidates based on our criteria (see Appendix~\ref{sec:inflated_prior_generation}). Each remaining segmentation mask, treated as a part $p$, is then converted into a 2D mesh by triangulating its contour. We inflate this mesh to produce a local height field $h_p$, yielding a \textbf{Detail 3D} component. We then superimpose this local height field on the base height by interpolating its values onto the base mesh vertices.

Repeating this process over all parts simply sums these localized offsets on the front height field: 
\begin{equation}
    h_{\text{final}} = h_{\text{base}} + \sum_p \mathcal{I}_p\big(h_p\big),
    \label{eq:part_emboss}
\end{equation}
where $h_{\text{final}}$ is computed by summing the global Base 3D height field ($h_{\text{base}}$) with the interpolated local height fields ($\mathcal{I}_p(h_p)$) from all parts $p$, where $\mathcal{I}_p$ denotes the piecewise-linear interpolation function. This final height field is then lifted to create the \textbf{Inflated Prior} mesh for Stage~2. In contrast, using a generic primitive such as a sphere as the prior provides no image-aligned cues and fails to generate voluminous, image-consistent geometry with detailed structures (see Appendix~\ref{sec:analysis_inflatedprior}).

\begin{figure}[t]
  \centering
  \includegraphics[width=1.0\linewidth]{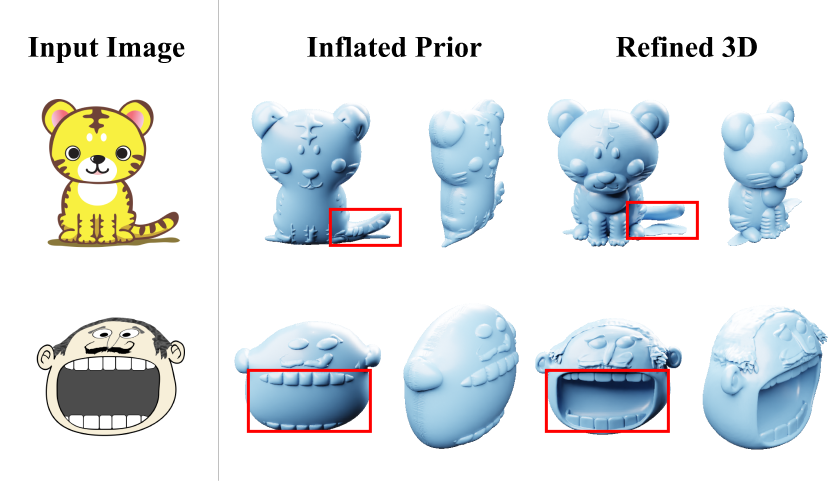}
  \caption{Stage~1's additive method incorrectly forms features that should be back-facing (like a tail) or concave (like a mouth) as simple convex regions, resulting in convex-only geometry.}
  \label{fig:details}
\end{figure}

\begin{figure*}[t!]
  \centering
  \includegraphics[width=1.0\textwidth]{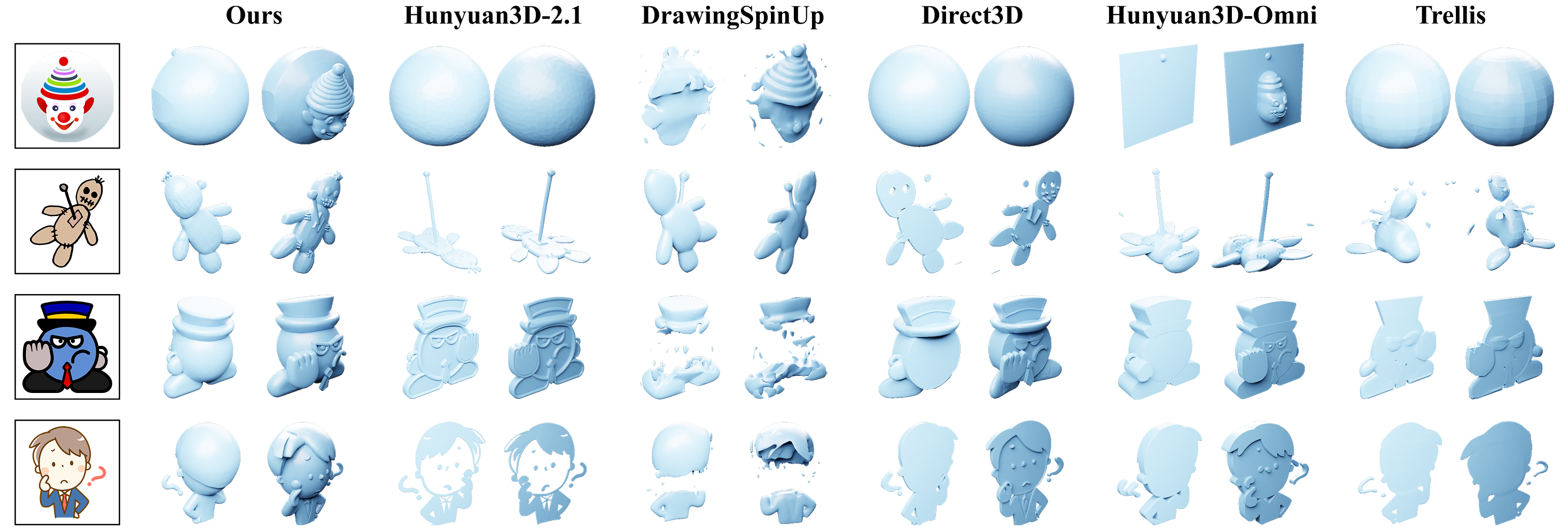}
  \caption{Visual comparison of our method against baseline methods, using the Hunyuan3D-2.1~\cite{hunyuan3d2025hunyuan3d} as the backbone.}
  \label{fig:qualitative_comparison}
\end{figure*}

\subsection{3D Latent Refinement}
\label{sec:refinement}
Although the Inflated Prior provides essential volumetric cues, its additive nature inevitably leads to \emph{convex-only geometry}. This means both back-facing structures and concave features are incorrectly formed as convex regions (Fig.~\ref{fig:details}). To refine this convex-only geometry, we inject Gaussian noise into the 3D latent to leverage the diffusion model’s learned prior during denoising. The diffusion transformer uses image conditioning guided by part-aware and volumetric cues from the Inflated Prior.

\paragraph{Backbone Architecture.}

Conditional 3D latent generative backbones are typically composed of three key components. \textbf{3D Encoder} compresses a 3D representation into a compact latent $z$. \textbf{Diffusion Transformer} (DiT) $\Phi$ implements a conditional diffusion process in the latent space: given a noisy latent $z_t$ at time $t$ and an image embedding $\psi(y)$, it predicts a less noisy latent, where $y$ is the input image and $\psi(\cdot)$ is a pretrained image encoder~\cite{radford2021learning,oquab2023dinov2}. By iteratively applying this diffusion process over a time sequence $t \in [0, 1]$, the backbone maps an initial noisy latent to a clean latent $z_0$. \textbf{3D Decoder} then maps the denoised latent to a neural field and reconstructs an explicit 3D mesh using Marching Cubes.

\paragraph{Stochastic Refinement.}
Motivated by stochastic diffusion models~\cite{song2020score, meng2021sdedit} and their ability to correct errors and improve sample quality~\cite{karras2022elucidating}, we leverage this behavior to refine the Inflated Prior.

Specifically, our refinement process begins by encoding the Inflated Prior into a latent $z_0$. We then initialize the process at a normalized initial noise level $t_0 \in [0, 1]$, where $t_0=0$ corresponds to the clean latent (the encoded data) and $t_0=1$ corresponds to pure Gaussian noise. We inject this noise into $z_0$ based on $t_0$ to obtain the starting latent $z_{t_0}$:
\begin{equation}
    z_{t_0} = a_{t_0} z_0 + b_{t_0} \varepsilon, \qquad \varepsilon\sim\mathcal{N}(0,\mathbf{I}).
    \label{eq:noising}
\end{equation}
Here, $a_{t_0}$ and $b_{t_0}$ are time-dependent scaling coefficients determined by the diffusion transformer’s noise schedule within the backbone. This Gaussian noise injection adds stochasticity to the refinement, enabling the diffusion model to leverage the backbone’s pretrained 3D knowledge to correct the convex-only geometry. The subsequent refinement then proceeds from $z_{t_0}$ to $z_0$ using the conditional diffusion step. Guided by detailed structural cues from the prior, the model generates precise local structures while preserving essential volumetric cues. Finally, this refined latent is passed to the 3D Decoder to generate the final high-fidelity, voluminous 3D mesh.
\section{Experiments}
\label{sec:experiments}

This section presents comprehensive qualitative and quantitative evaluations against diverse baselines. The evaluations indicate strong semantic alignment with the input images and utilize our proposed metrics to demonstrate improved volume quantification. This improved quantification is further validated by a user study, which indicates alignment with human perception of volume and details. 

\paragraph{Dataset.}
Publicly available datasets for flat images remain limited in size and diversity. For example, the Art3D dataset~\cite{cong2025art3d} contains only about 100 images. To enable a robust evaluation, we collected a new test set of 2,232 flat images across diverse subjects such as humans, animals, and characters. This test set is curated to include cases where existing 3D generation models typically fail to produce voluminous geometry, such as flat-color images with almost no shading or shadows, and images where the subject is partially occluded or overlaps with the background.

\paragraph{Implementation Details.}
We implement our method using the Hunyuan3D-2.1~\cite{hunyuan3d2025hunyuan3d} and Direct3D~\cite{wu2024direct3d} backbones. In Stage~1, we set the inflation strength to $c=1.5$ for both the global contour and all local parts. For Stage~2, we use a guidance scale set to $7.0$ and $50$ sampling steps, and set the initial noise level to $t_0 = 0.8$ by default. Inference requires roughly 3 minutes per image (2 minutes for Stage~1 and 1 minute for Stage~2) on an NVIDIA RTX 6000 Ada GPU.

\subsection{Evaluation Metrics}
Evaluating the volume and surface flatness of 3D meshes generated from flat images is challenging because ground-truth 3D shapes are unavailable. As a result, reference-dependent metrics such as Chamfer Distance are not directly applicable. We introduce two metrics, Compactness and Normal Anisotropy, which quantify the volume and surface flatness of the generated meshes.

\paragraph{Compactness.}
Compactness \(C\) measures the volumetric compactness of a shape. Following the isoperimetric quotient, it is defined as the scale-invariant ratio of the squared volume ($V^2$) to the cubed surface area ($S^3$):
\begin{equation}
    C = \frac{36\pi V^2}{S^3}.
    \label{eq:compactness}
\end{equation}
This metric is normalized so that its range is $C \in [0, 1]$. To robustly handle non-watertight meshes, we voxelize the mesh into a solid occupancy and regularize it via standard volumetric morphology (see Appendix~\ref{sec:metrics}). We utilize a fixed voxel pitch of $0.02$ when processing the voxel grid. These steps ensure a consistent and robust measurement of $V$ and $S$ across all generated meshes.

\paragraph{Normal Anisotropy.}
However, Compactness alone is insufficient to assess the geometric quality of the generated meshes with respect to the input image. A simple shape such as an inflated eraser can achieve a high $C$ by increasing its volume, yet it exhibits high surface flatness and lacks local structure, resulting in poor correspondence to the input image. To quantify this surface flatness, which is characterized by a low-entropy normal distribution, we propose \emph{Normal Anisotropy} (NA). This metric quantifies the anisotropy of the mesh's area-weighted face-normal distribution. We discretize the sphere of normal directions into \(K=128\) uniform bins and compute a probability \(p_k\) by summing normalized face areas in each bin. NA is defined as the normalized complement of the distribution's Shannon entropy:
\begin{equation}
    \mathrm{NA}(\mathcal{M}) = 1 - \frac{-\sum_{k=1}^{K} p_k \log(p_k + \epsilon)}{\log K},
    \label{eq:na}
\end{equation}
where $\mathrm{NA}\in[0,1]$. Flat or anisotropic geometry concentrates face normals in a few directions, resulting in low entropy and a high $\mathrm{NA}$.

\begin{table}[t!]
  \centering
  \caption{Category ranking on ModelNet40~\cite{wu20153d} by Compactness and Normal Anisotropy. \textbf{Left}: top three and bottom three by Compactness. \textbf{Right}: top three and bottom three by Normal Anisotropy}
  \label{tab:modelnet_results}
  \setlength{\tabcolsep}{3pt} 
  \begin{tabular}{@{} l c | l c @{}}
      \toprule
      \multicolumn{2}{c |}{Compactness} & \multicolumn{2}{c}{Normal Anisotropy} \\
      \midrule
      glass\_box & 0.3983 & door     & 0.6591 \\
      dresser    & 0.3669 & wardrobe & 0.5916 \\
      radio      & 0.3237 & keyboard & 0.5516 \\
      \midrule 
      curtain    & 0.0137 & vase     & 0.1364 \\
      keyboard   & 0.0175 & bowl     & 0.1525 \\
      laptop     & 0.0270 & plant    & 0.1592 \\
      \bottomrule
  \end{tabular}
\end{table}

\begin{table}[t!]
  \small
  \caption{Quantitative comparisons of our method against baselines, evaluated using Uni3D, ULIP, Compactness (\(C\)), and Normal Anisotropy ($\mathrm{NA}$).}
  \label{tab:quantitative_results}
  \setlength{\tabcolsep}{0pt}
  
  \begin{tabularx}{\linewidth}{@{}c | C C C C@{}}
      \toprule
      
      Models & Uni3D$\uparrow$ & ULIP$\uparrow$ & \(C\)$\uparrow$ & $\mathrm{NA}$$\downarrow$ \\
      \midrule
      Trellis~\cite{xiang2024structured}            & 0.2736 & 0.1241 & 0.1748 & 0.1282 \\
      DrawingSpinUp~\cite{zhou2024drawingspinup}     & 0.2335 & 0.1164 & 0.1604 & 0.1332 \\
      Hunyuan3D-Omni~\cite{hunyuan3d2025hunyuan3domni}   & 0.2816 & 0.1257 & 0.1707 & 0.1120 \\
      Direct3D~\cite{wu2024direct3d}          & 0.2796 & 0.1315 & 0.2012 & 0.1019 \\
      Hunyuan3D-2.1~\cite{hunyuan3d2025hunyuan3d}    & 0.2759 & 0.1193 & 0.1408 & 0.1347 \\
      \midrule 
      Ours(Hunyuan3D-2.1) & 0.3043 & 0.1265 & \textbf{0.2179} & \textbf{0.0767} \\
      Ours(Direct3D)    & \textbf{0.3097} & \textbf{0.1375} & 0.2178 & 0.0908 \\
      \bottomrule
  \end{tabularx}
\end{table}

\begin{figure}[t]
  \centering
  \includegraphics[width=1.0\linewidth]{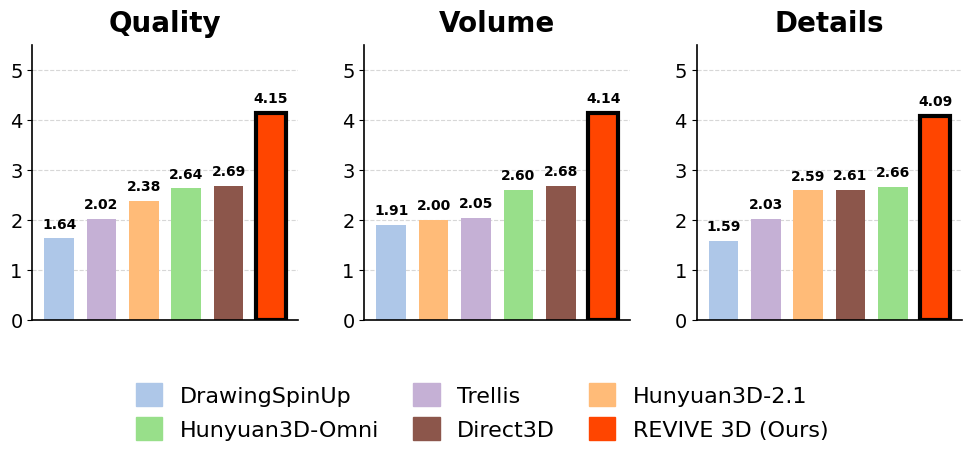}
  \caption{User study ratings on a 5-point Likert scale for Quality, Volume, and Details.}
  \label{fig:userstudy}
\end{figure}

\paragraph{Metric Validation.}
To validate our metrics, we ranked ModelNet40~\cite{wu20153d} categories by \(C\) and \(\mathrm{NA}\) and inspected the top three and bottom three. The results (Table~\ref{tab:modelnet_results}) align with intuition. For \(C\), volumetric categories (\texttt{glass\_box}, \texttt{dresser}, \texttt{radio}) scored highest, while flat categories (\texttt{curtain}, \texttt{keyboard}, \texttt{laptop}) scored lowest. For \(\mathrm{NA}\), flat categories (\texttt{door}, \texttt{wardrobe}, \texttt{keyboard}) scored highest, while categories with complex, curved surfaces (\texttt{vase}, \texttt{bowl}, \texttt{plant}) scored lowest. These observations suggest that \(C\) correlates with volume and \(\mathrm{NA}\) with surface flatness. Therefore, a high $C$ indicates voluminous 3D meshes, and a low $\mathrm{NA}$ indicates low surface flatness.

\subsection{Generation Results}
\paragraph{Qualitative Comparisons.}
We conduct qualitative comparisons (Fig.~\ref{fig:qualitative_comparison}) against the structured 3D generator Trellis~\cite{xiang2024structured}, the character-specialized DrawingSpinUp~\cite{zhou2024drawingspinup}, and Hunyuan3D-Omni with bounding-box conditioning~\cite{hunyuan3d2025hunyuan3domni}. We also compare against the backbone-only baselines Direct3D~\cite{wu2024direct3d} and Hunyuan3D-2.1~\cite{hunyuan3d2025hunyuan3d} without our method.
Across baselines, meshes often lack global volume, fine detail, or both. DrawingSpinUp shows high variability across inputs. Hunyuan3D-Omni increases apparent volume with bounding-box conditioning but often stretches or inflates specific parts to meet the box, producing uniform volumetric expansion that breaks image consistency. Our method generates image-consistent volume and fine detail from flat images and outperforms all baselines.

\begin{figure}[t]
  \centering
  \includegraphics[width=1.0\linewidth]{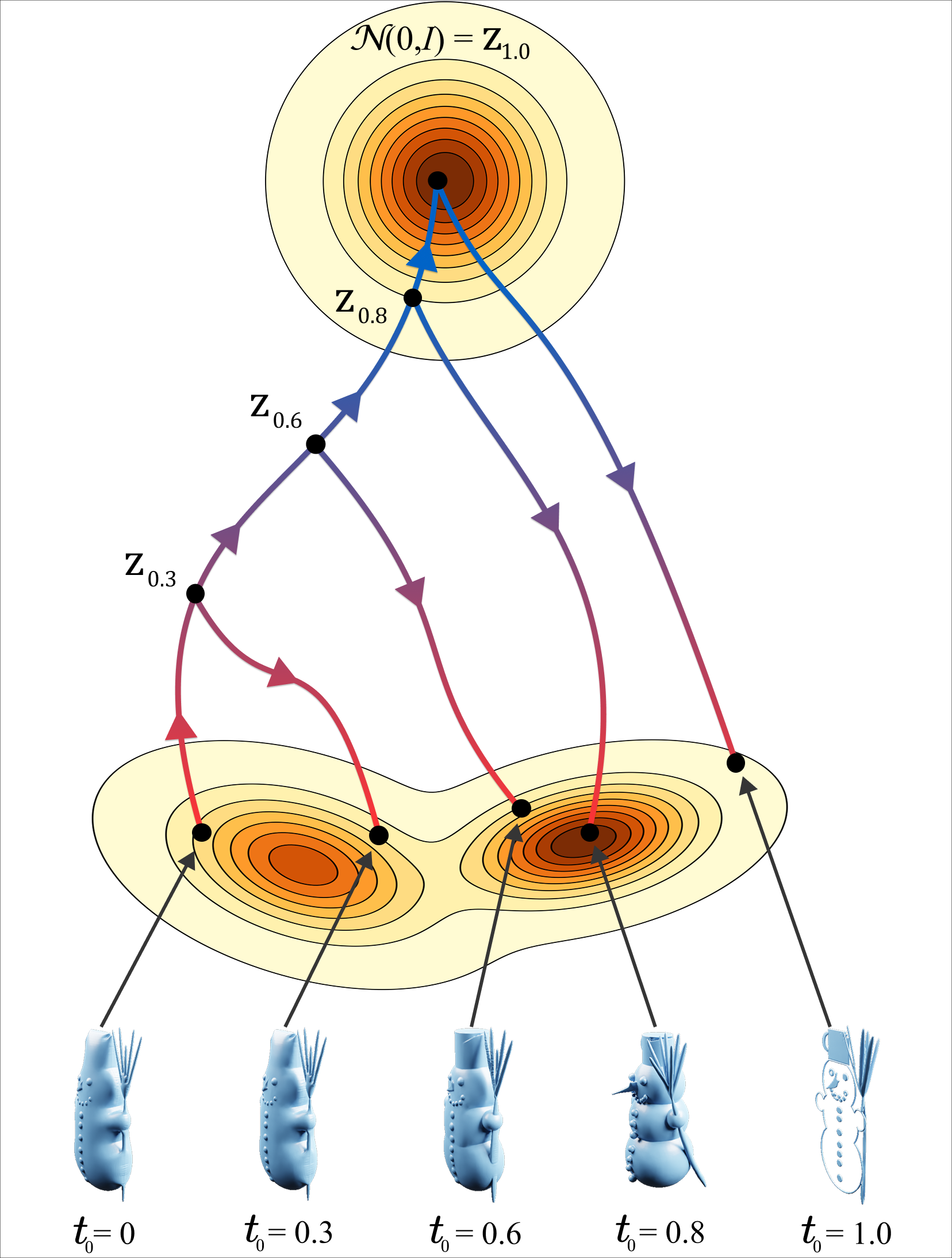}
  \caption{Refinement trajectories for different initial noise levels \(t_0\). The circle at the top visualizes pure Gaussian noise with arrows pointing toward the naive backbone result on the right. Larger \(t_0\) starts closer to this region and moves away from the Inflated Prior, whereas smaller \(t_0\) keeps the trajectory near the Inflated Prior.}
  \label{fig:skip}
\end{figure}

\begin{figure}[t]
  \centering
  \includegraphics[width=1.0\linewidth]{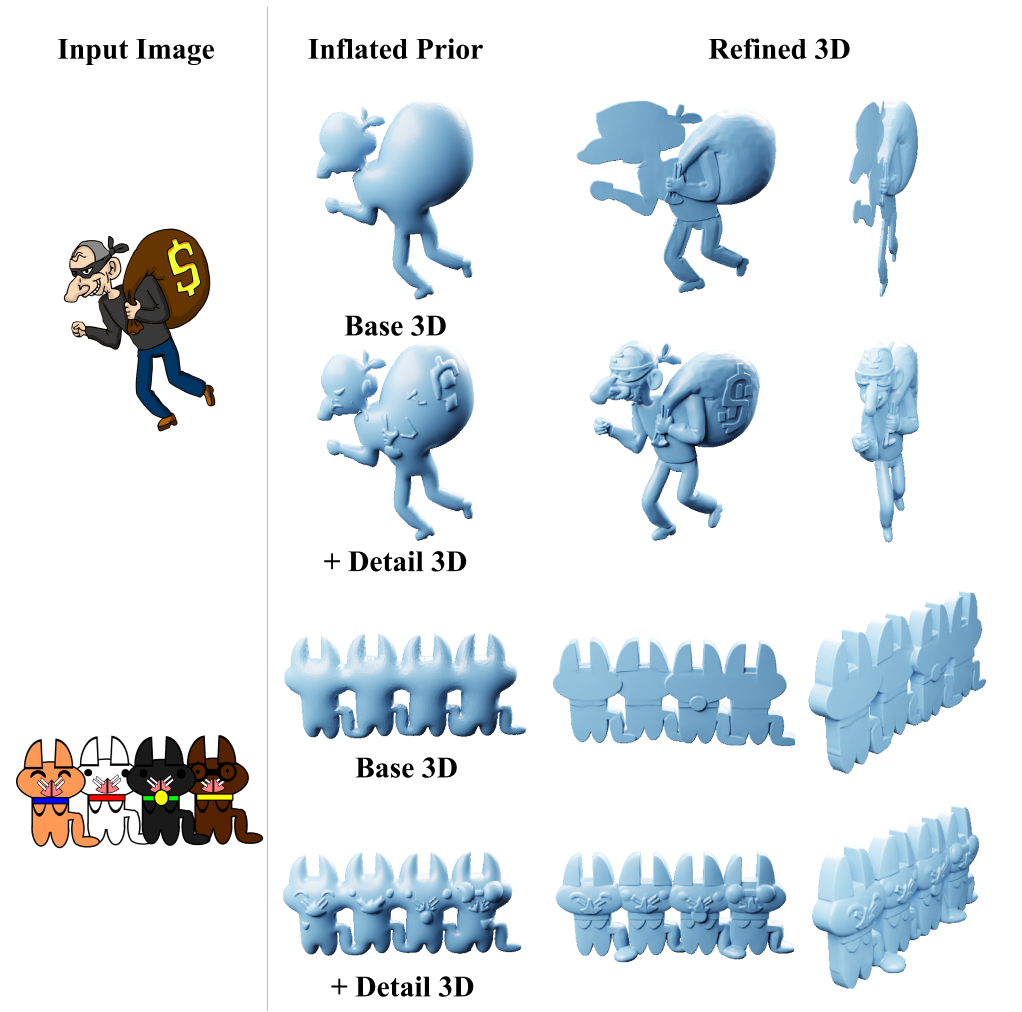}
  \caption{Each row shows a different example. Within each example, the top row displays the refined result using only the Base 3D, which struggles to generate a voluminous and detailed mesh. The bottom row shows the result using the complete Inflated Prior with the Detail 3D, which successfully generates a 3D mesh with voluminous global shape and rich local details.}
  \label{fig:superimposing}
\end{figure}


\paragraph{Quantitative Comparisons.}
For a comprehensive quantitative comparison, we use our proposed metrics, Compactness and Normal Anisotropy, to measure volume and surface flatness. Additionally, to evaluate image–3D consistency, we compute cross-modal similarity scores using both ULIP~\cite{xue2023ulip} and Uni3D~\cite{zhou2023uni3d}. These models are specifically designed to learn a unified feature space by aligning 3D point cloud representations with image and text features. We therefore leverage their powerful encoders to extract a feature embedding from the input image and another from the generated 3D mesh. A high cosine similarity between these embeddings signifies strong semantic alignment between the generated 3D mesh and the input image.

As shown in Table~\ref{tab:quantitative_results}, our method outperforms all baselines. This demonstrates that our method produces shapes with high volume (high \(C\)) and low surface flatness (low \(\mathrm{NA}\)). It also attains higher ULIP and Uni3D scores, signifying our method’s strong semantic consistency with the input images.

\paragraph{User Study.} To supplement our quantitative metrics, we conducted a user study with 51 participants to evaluate the perceptual quality of our method against the 5 baselines. We randomly selected 10 challenging images from our test set. For each image, participants were shown auto-rotating GIFs (360-degree renders) of the 3D models from all 6 methods in a randomized order and asked to rate them on a 5-point Likert scale based on the following three questions:

\begin{itemize} \item \textbf{Quality.} How satisfied are you with the overall visual quality of the 3D model? \item \textbf{Volume.} How convincing is the 3D model's volume and 3-dimensional shape (does it look flat)? \item \textbf{Details.} How clear and accurate are the model's fine details (surface features, patterns)? \end{itemize}

\begin{figure*}[t!]
  \centering
  \includegraphics[width=\linewidth]{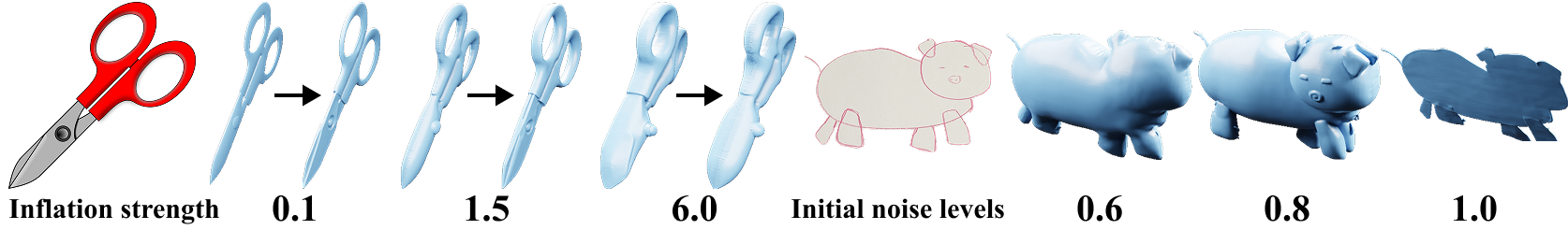}
  \caption{Hyperparameter robustness analysis. Arrows indicate the direction from the prior to the final 3D results.}
  \label{figure2}
\end{figure*}

As shown in Fig.~\ref{fig:userstudy}, the results indicate that our method was the most preferred across all categories, achieving the highest scores for quality, volume, and details.

\subsection{Ablation Study}
\paragraph{Effect of initial noise level $t_0$.}
The initial noise level $t_0$ controls a fidelity–plausibility trade-off in our refinement process. In our setting, this trade-off directly affects the resulting volume (Fig.~\ref{fig:skip}). A low $t_0$ initializes $z_{t_0}$ close to the Inflated Prior, retaining high fidelity and preserving its volume but also its convex-only geometry. Conversely, a high $t_0$ moves $z_{t_0}$ toward pure Gaussian noise, so denoising by the diffusion model leverages the backbone’s pretrained 3D knowledge to correct convex-only geometry while reducing the volume provided by the Inflated Prior.

We choose the initial noise level \(t_0\) to preserve the Inflated Prior’s injected volume while enabling the diffusion model’s stochasticity to refine its convex-only geometry. Empirically, we find that setting \(t_0\) in the range \([0.7, 0.8]\) enables the diffusion model to correct convex-only geometry while retaining the prior’s volumetric cues.

\paragraph{Benefits of Local Part Superimposing.}
Local Part Superimposing in Stage~1 is essential for effective refinement. An Inflated Prior derived from Global Contour Inflation lacks localized geometric cues, making it difficult to refine local structure. As shown in Fig.~\ref{fig:superimposing}, this often yields meshes with high surface flatness or loss of details even when global volume is recovered. In contrast, we construct a part-aware Inflated Prior by superimposing the part contours, thereby providing explicit cues about the locations and shapes of fine-grained parts to guide refinement. The conditional diffusion transformer then uses these cues to better align the image features with the latent geometry, producing deformations that respect both the Inflated Prior’s global volumetric structure and image-aligned local details.

\begin{table}[t]
\centering
\small
\setlength{\tabcolsep}{2pt}
\renewcommand{\arraystretch}{1.0}
\begin{tabular}{c c
                S[table-format=1.4]
                S[table-format=1.4]
                S[table-format=1.4]
                S[table-format=1.4]}
\hline
Inflation & Noise & {C} & {NA} & {Uni3D} & {ULIP} \\
\hline
6.0 & \textcolor{red}{0.8} & {0.2682} & {0.0547} & {0.3276} & {\bfseries 0.1153} \\
0.1 & \textcolor{red}{0.8} & 0.1172 & 0.2539 & 0.2840 & 0.0935 \\
\textcolor{red}{1.5} & 1.0 & 0.1501 & 0.2168 & 0.3003 & 0.1006 \\
\textcolor{red}{1.5} & 0.6 & {\bfseries 0.4296} & {\bfseries 0.0511} & 0.3043 & 0.0979 \\
\textcolor{red}{1.5} & \textcolor{red}{0.8} & 0.2691 & 0.0610 & {\bfseries 0.3382} & {\bfseries 0.1153} \\
\hline
\end{tabular}
\vspace{-1ex}
\caption{Ablation over inflation strength (shared for global and local) and initial noise levels. Default setting is highlighted in red.}
\label{tab:ablation}
\end{table}

\paragraph{Hyperparameter robustness.}
We analyze the hyperparameter robustness in Fig.~\ref{figure2} by varying inflation strength and initial noise level. Quantitative results in Table~\ref{tab:ablation} on the Art3D~\cite{cong2025art3d} dataset support consistent behavior. At noise 0.6, the result is closer to the inflated prior (C, NA), but noise 0.8 better aligns geometry with the input image (Uni3D/ULIP), which we choose as default.
\begin{figure}[t!]
  \centering
  \includegraphics[width=\linewidth]{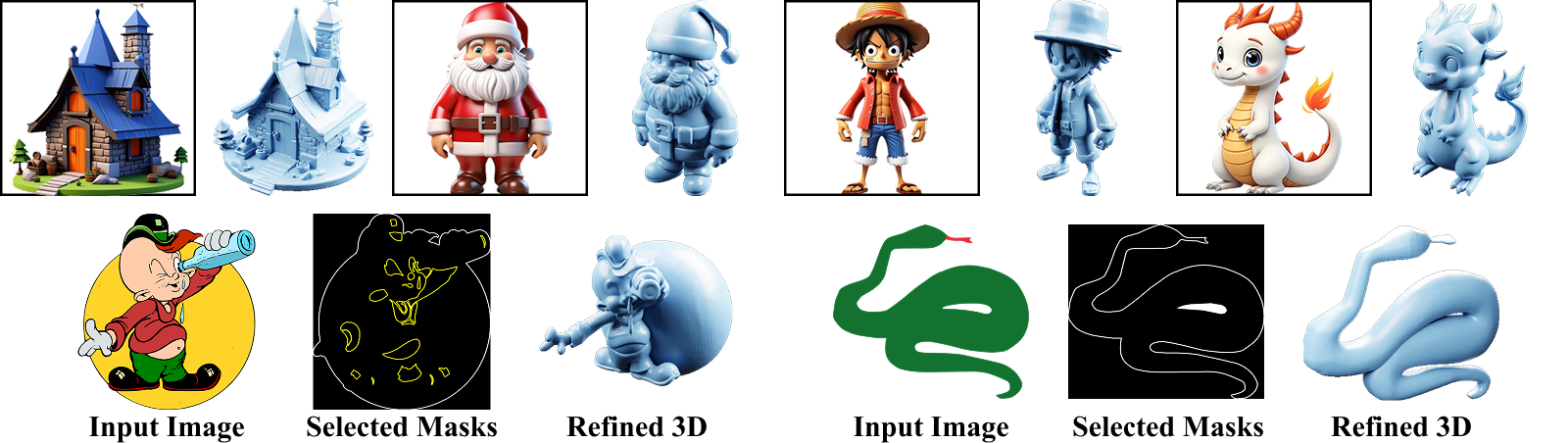}
  \caption{Qualitative results showing non-flat cases (top) and challenging cases with ambiguous silhouettes and imperfect segmentation (bottom; white: silhouette mask, yellow: segmentation mask).}
  \label{figure1}
\end{figure}

\subsection{Generalization}
While our method targets a specific failure mode, it does not impose additional input constraints beyond those already required by the pretrained backbone. As shown in Fig.~\ref{figure1}, the bottom panel demonstrates robustness to ambiguous silhouettes and imperfect segmentation, while the top panel shows applicability to non-flat inputs.

\section{Conclusion}
We presented REVIVE 3D, a two-stage, plug-and-play pipeline that generates voluminous 3D meshes from flat images. We constructed an Inflated Prior by inflating the outer contour and superimposing part-aware inflations, supplying the missing volumetric and part-level cues. The subsequent 3D Latent Refinement injects Gaussian noise at a specific initial noise level to overcome the prior's convex-only geometry and leverage the backbone’s pretrained 3D knowledge. 
Furthermore, we introduced Compactness and Normal Anisotropy to quantify the volume and surface flatness of the generated geometry.

A limitation of our approach is its reliance on the pretrained backbone for refinement, which can shift the stylistic simplicity of cartoon inputs toward the backbone’s photorealistic bias. We expect this shift to be mitigated by better texture alignment with the input image, which we leave for future work. Nevertheless, extensive qualitative and quantitative comparisons show that REVIVE 3D generates voluminous and detailed 3D meshes from flat images.
\section{Acknowledgement}
This work was supported by the National Research Foundation of Korea (NRF) grant funded by the Korea government (MSIT) (RS-2025-02216328), by the Culture, Sports and Tourism R\&D Program through the Korea Creative Content Agency grant funded by the Ministry of Culture, Sports and Tourism in 2024 (RS-2024-00398413; Contribution Rate: 50\%), by the High-Performance Computing Support Project, funded by the Government of the Republic of Korea (Ministry of Science and ICT) (RQT-25-070274), and by the Korea Technology and Information Promotion Agency for SMEs (TIPA) (grant funded by the Ministry of SMEs and Startups (MSS)) (No. RS-2024-00446233). We thank Minjae Kang for valuable feedback and insightful discussions and Yuna Shin for generous help and support.
{
    \small
    \bibliographystyle{ieeenat_fullname}
    \bibliography{main}
}
\clearpage
\twocolumn[{ 
    \begin{center}
        \textbf{\Large REVIVE 3D: Refinement via Encoded Voluminous Inflated prior for Volume Enhancement} \\
        \vspace{0.3cm}
        \Large (Supplementary Material)
    \end{center}
    \vspace{0.8cm} 
}]
\setcounter{section}{0} 
\renewcommand{\thesection}{\Alph{section}} 
\renewcommand{\thesubsection}{\Alph{section}.\arabic{subsection}} 

\begin{figure}[t]
  \centering
  \includegraphics[width=1.0\linewidth]{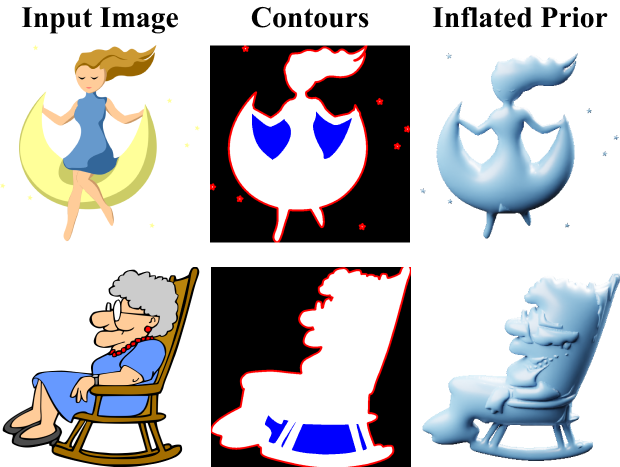}
  \caption{Contour and Cut Visualization. The outer boundary is marked in red, and the excluded internal regions are filled in blue.}
  \label{fig:conturs_cut}
\end{figure}

\begin{figure*}[t!]
  \centering
  \includegraphics[width=1.0\textwidth]{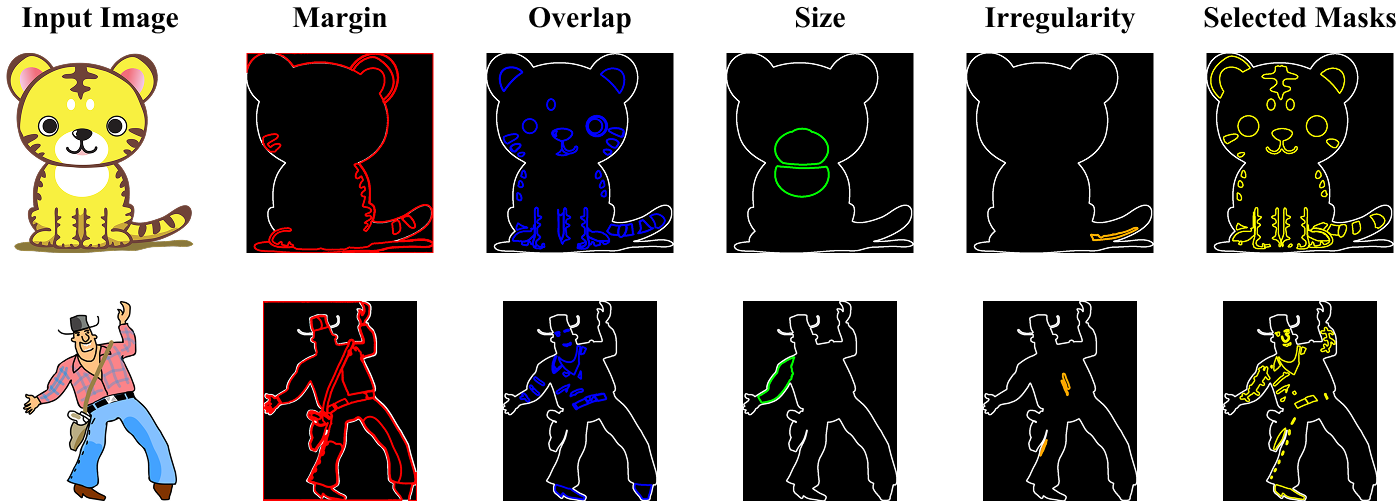}
  \caption{Visualization of the mask filtering process. We visualize masks filtered by our four criteria: Margin (red), Overlap (blue), Size (green), and Irregularity (orange). The final column shows the aggregated selected masks (yellow) that pass all filters.}
  \label{fig:mask_filter}
\end{figure*}

\section{More Implementation Details} 
\label{sec:more_details}

\subsection{Generation of the Inflated Prior}
\label{sec:inflated_prior_generation}
We automatically generate an Inflated Prior to serve as a volumetric and part-aware prior for Stage~2. This is done by combining a global silhouette mask with local part masks derived from an automatic segmentation method.

The silhouette mask is a binary mask generated from the input image, designating the foreground object area as 255 and the transparent background as 0. The boundaries of the foreground regions define one or more outer contours. Simply inflating these outer contours as a single solid would cause interior empty regions to be filled and disconnected parts to be merged, leading to a loss of the original overall shape. To prevent this, we explicitly exclude these internal empty areas from the inflation process (Fig.~\ref{fig:conturs_cut}).

To obtain part-aware local regions, one could in principle apply traditional edge detection. However, standard edge detectors struggle to accurately capture subtle boundaries and often fail to produce precise, closed contours that are usable for our inflation process. Therefore, instead of edges, we rely on an automatic segmentation-based approach~\cite{ravi2024sam2} to extract local parts from the input image. The resulting segmentation masks vary greatly in size, location, and shape, so we apply a filtering process based on four specific criteria (Fig.~\ref{fig:mask_filter}).

First, we filter based on \textbf{Margin}. We only keep masks that lie completely inside the main silhouette and discard any mask that extends outside the silhouette boundary. Masks that lie too close to the silhouette contour are also problematic, because the height field is constrained to zero along the contour. Inflating such masks can introduce rendering artifacts or distort the final outline. To avoid this, we remove any mask judged to be too close to a boundary. Concretely, a mask is removed if it is located within 6 pixels of the main silhouette contour, if it overlaps with the 3-pixel eroded region of the contour, or if it is within 16 pixels of an internal empty region. Second, we eliminate \textbf{Overlap}. We sort all masks by size in descending order to establish priority. Any smaller mask that overlaps with an already selected larger mask is discarded. While smaller masks can represent fine details, we prioritize the larger, more dominant cues for providing clear guidance in Stage~2. Third, we apply a \textbf{Size} constraint. Masks smaller than 16 pixels are removed as they are typically noise or are difficult to sample. Conversely, masks larger than 10,000 pixels are also removed because they can overwhelm and obscure other useful detail cues. Finally, we correct for \textbf{Irregularity}. Unusually thin masks with an aspect ratio exceeding 1:5 are removed, as such thread-like regions tend not to be refined and often appear as distracting artifacts in the final mesh. We also use a 3-pixel erosion test to assess the mask's thickness. Line-like masks that lose over 95 percent of their original area upon erosion are discarded. Masks that lose between 60 and 95 percent of their area are considered viable but thin, so they are reinforced with a 5-pixel dilation and then used.

The resulting set of filtered masks is then superimposed onto the global silhouette. This process creates the final Inflated Prior, which provides the necessary detailed part cues for Stage~2.

\begin{figure}[t]
  \centering
  \includegraphics[width=1.0\linewidth]{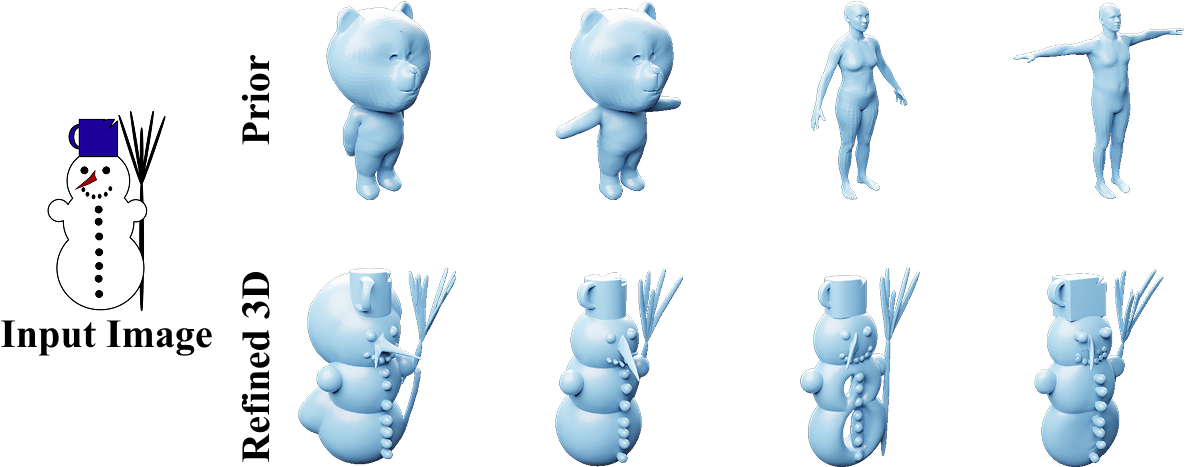}
  \caption{Prior and refined 3D results when using character and human body template meshes as priors.}
  \label{fig:all_prior}
\end{figure}

\begin{figure*}[t!]
  \centering
  \includegraphics[width=1.0\textwidth]{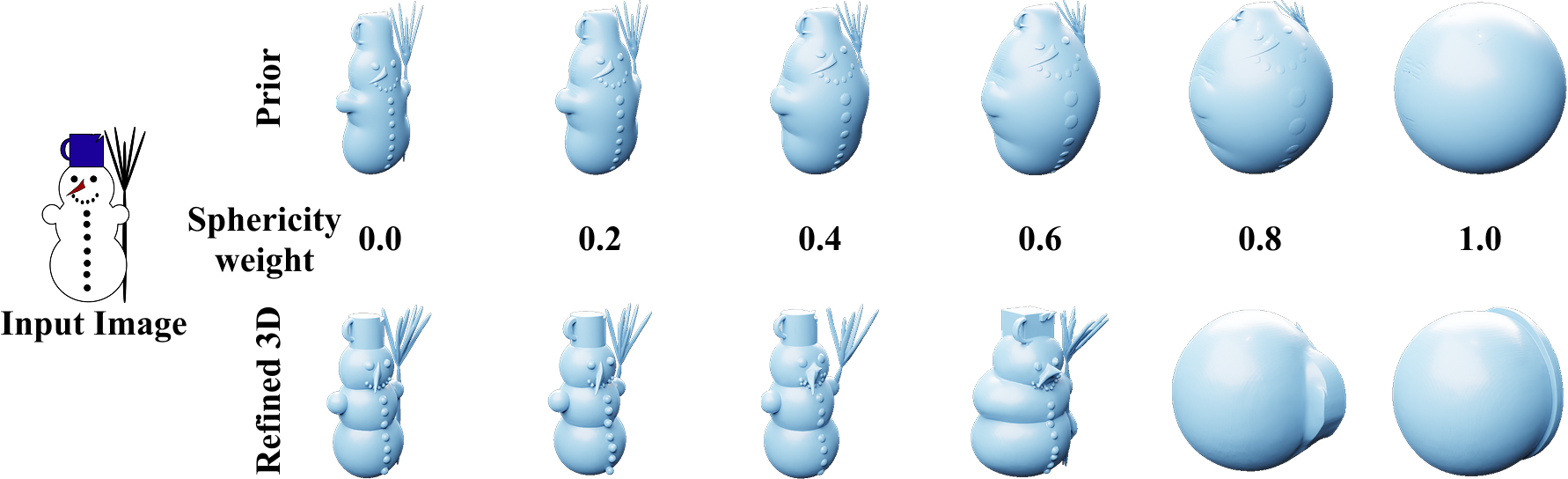}
  \caption{Interpolating the Inflated Prior toward a sphere degrades image-consistent geometry, as shown by priors (top) and refined 3D results (bottom) for different Sphericity weights from 0.0 to 1.0.}
  \label{fig:linear_prior}
\end{figure*}

\begin{figure*}[t!]
  \centering
  \includegraphics[width=1.0\textwidth]{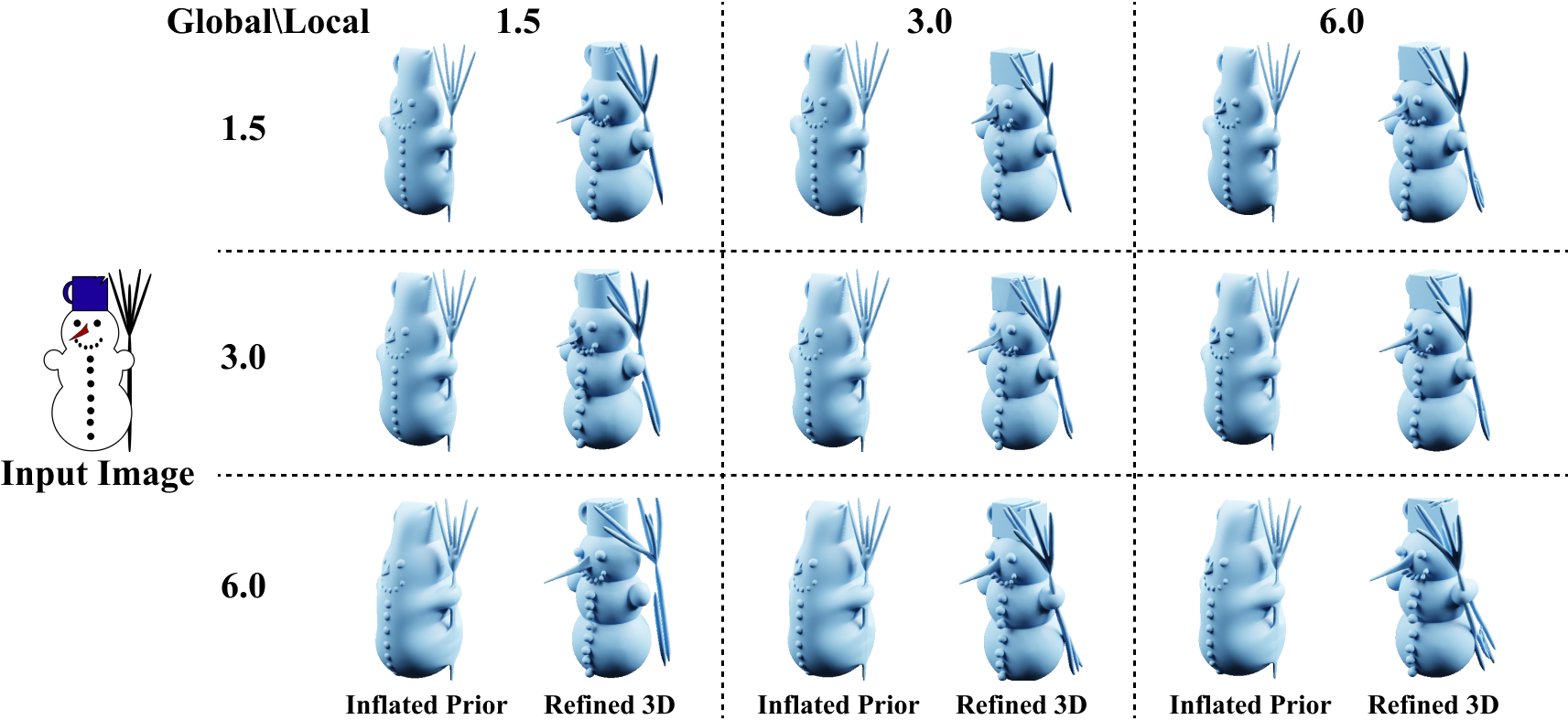}
  \caption{Effect of varying global and local inflation strength (1.5, 3.0, 6.0) on the Inflated Prior and refined 3D results.}
  \label{fig:strength}
\end{figure*}

\begin{figure*}[t!]
  \centering
  \includegraphics[width=1.0\textwidth]{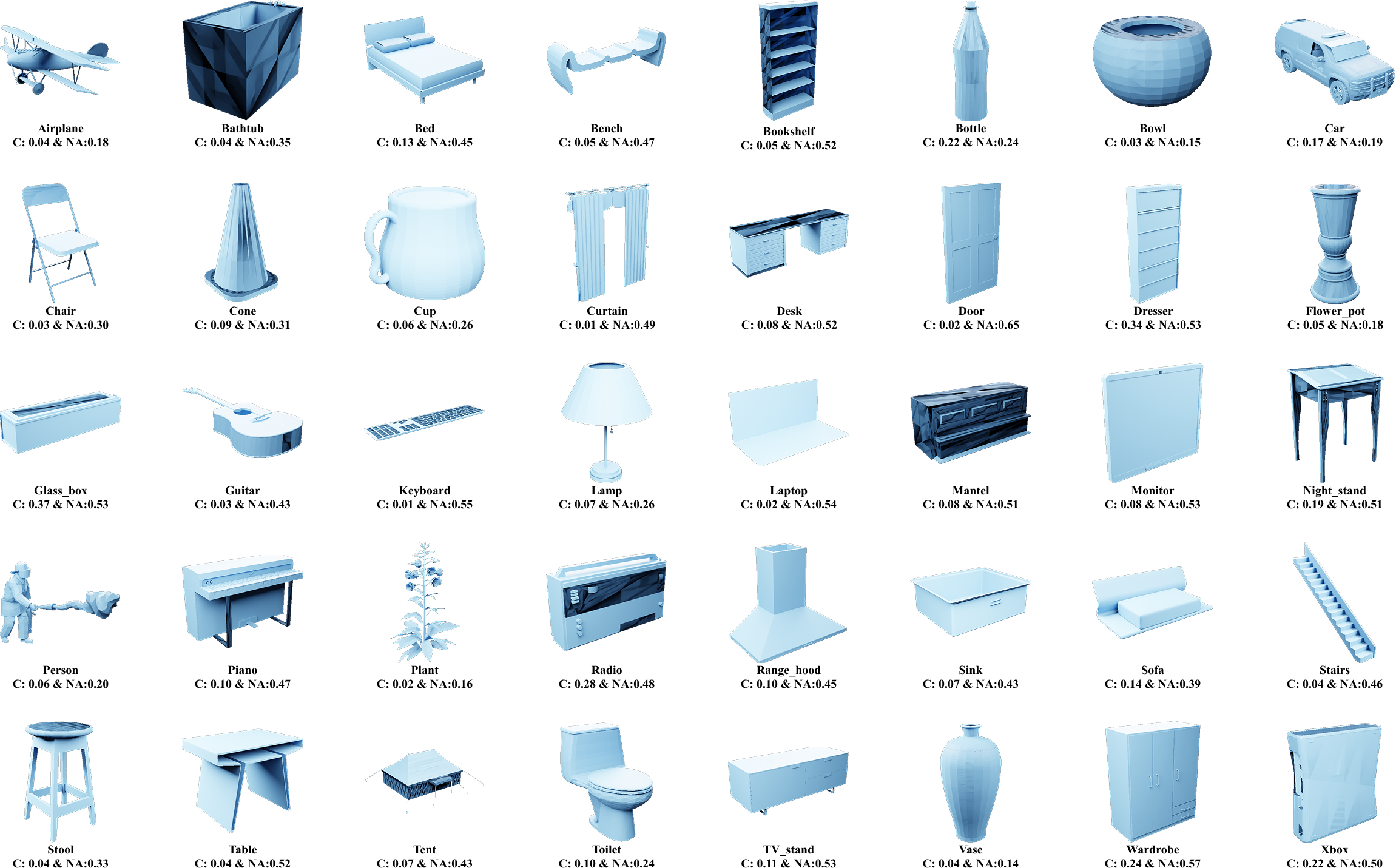}
  \caption{Category-wise Compactness (C) and Normal Anisotropy (NA) on ModelNet40~\cite{wu20153d}.}
  \label{fig:metric}
\end{figure*}

\begin{figure}[t]
  \centering
  \includegraphics[width=1.0\linewidth]{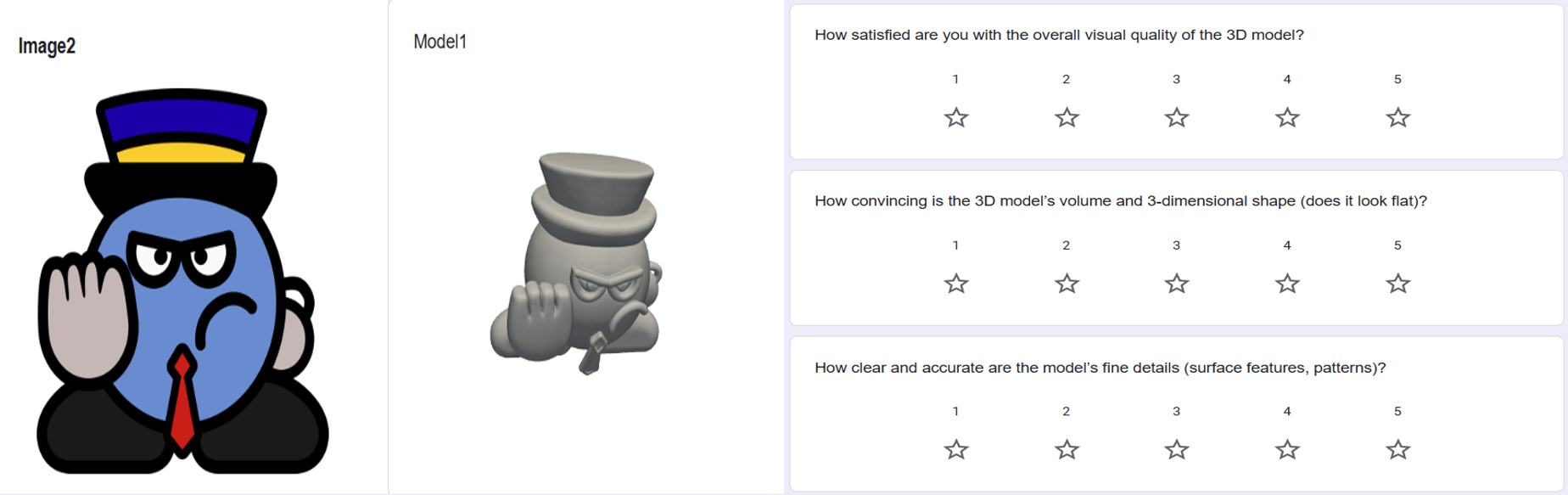}
  \caption{Example of the web interface used in our user study.}
  \label{fig:userstudy_interface}
\end{figure}

\begin{figure*}[t!]
  \centering
  \includegraphics[width=1.0\textwidth]{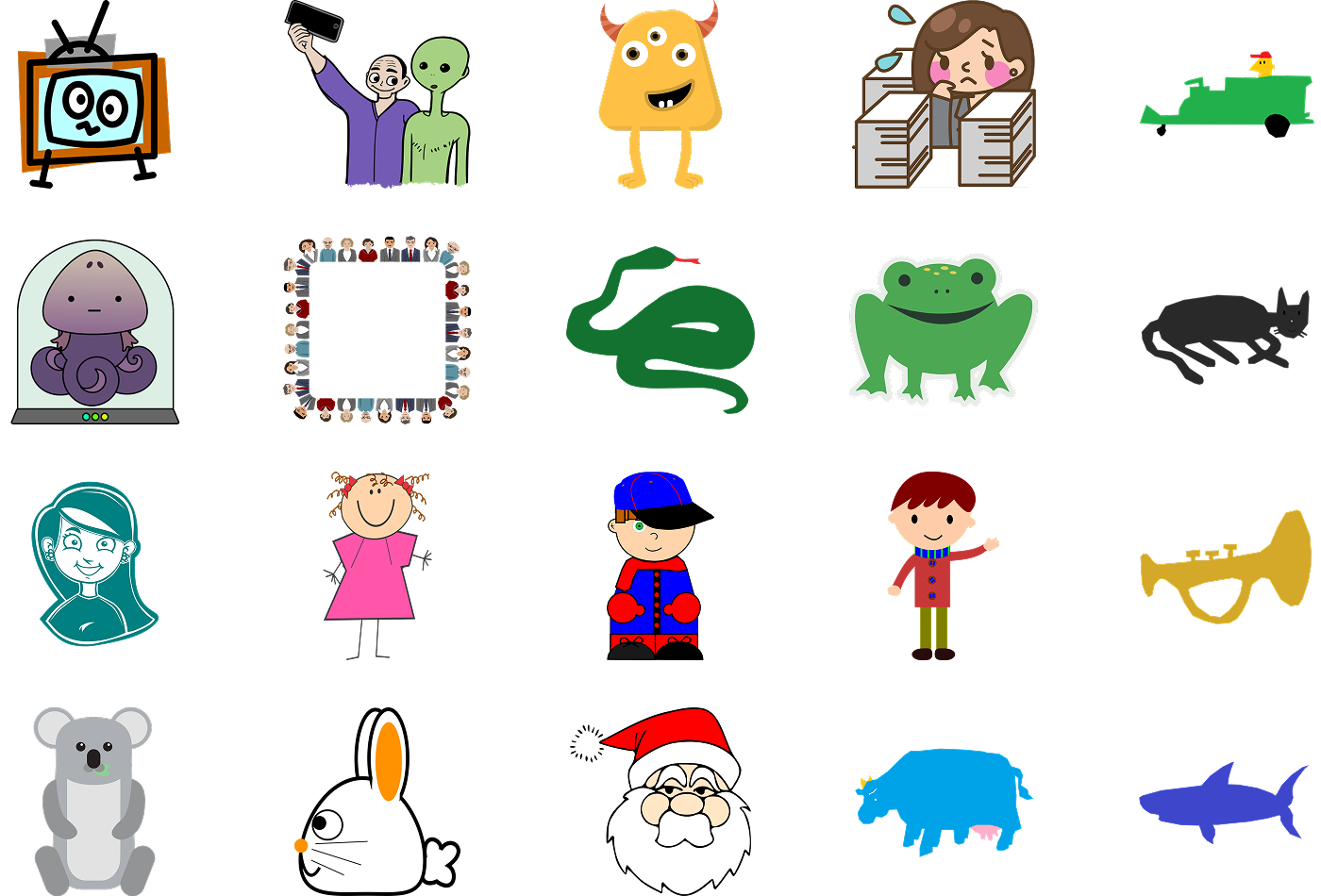}
  \caption{Dataset examples.}
  \label{fig:data_explain}
\end{figure*}

\subsection{Analysis of Inflated Prior}
\label{sec:analysis_inflatedprior}
\paragraph{Effect of Prior Specificity.}
To demonstrate the importance of the volumetric and part-aware cues provided by the proposed Inflated Prior when generating 3D meshes from flat images, we perform a deformation analysis. Stage~1 of our framework uses the input image silhouette and segmentation masks to construct an Inflated Prior that captures image-specific volume and local details. If this image-specific prior is replaced by a generic 3D mesh (Fig.~\ref{fig:all_prior}), the subsequent 3D Latent Refinement stage cannot fully reflect the structures in the input image and instead produces results that rely on the generic prior shape.

To quantify how this prior specificity influences the final output, we introduce an interpolation weight called \emph{Sphericity weight} (Fig.~\ref{fig:linear_prior}). Sphericity weight linearly interpolates vertex positions between the original Inflated Prior and a sphere, and therefore controls how close the prior shape is to the original image-specific Inflated Prior or to a sphere. When Sphericity weight is 0.0, the vertex coordinates of the original Inflated Prior are used without modification, which corresponds to the most image-specific setting. When Sphericity weight is 1.0, all vertices are normalized so that they lie at an equal distance from the center, resulting in a perfect sphere. As Sphericity weight increases from 0.0 to 1.0, the prior gradually loses image-specific cues and converges toward the shape of a sphere. This analysis confirms that constructing the Inflated Prior in an image-specific form plays a crucial role in the quality of the final 3D generation.

\paragraph{Effect of Inflation Strength $c$.}
We also analyze how the inflation strength value $c$, used when inflating the silhouette and the filtered masks, affects both the initial Inflated Prior and the final generated 3D mesh. In our framework, the global inflation strength applied to the silhouette and the local inflation strength applied to the segmentation masks can be controlled independently. Using the default setting $c$ equal to 1.5 as a reference, we conduct comparison experiments by increasing each of the global and local values to 3 and 6, which correspond to two and four times the default strength.

The results (Fig.~\ref{fig:strength}) show that increasing the global inflation strength causes large-scale structures to become overly inflated, which distorts and eventually breaks the overall silhouette. In contrast, increasing the local inflation strength exaggerates fine-scale features, leading to results where small details are emphasized. Based on these observations, we set the default inflation strength to $c$ equal to 1.5 for both global and local inflation in our experiments.

\subsection{Compactness and Normal Anisotropy}
\label{sec:metrics}

\paragraph{Mesh preprocessing.}
Before computing the metrics, we normalize each input mesh. If a file contains a scene with multiple mesh instances, we first merge all geometry into a single triangular mesh and remove unreferenced vertices and degenerate faces. We then translate the mesh so that its centroid is at the origin and rescale
it so that the longest side of its oriented bounding box has unit length. This preprocessing makes Compactness and Normal Anisotropy invariant to global translation and scale and reduces numerical issues caused by extreme coordinate ranges.

\paragraph{Compactness.}
For Compactness we require reliable estimates of the enclosed volume $V$ and surface area $S$ even for imperfect meshes. If a mesh is reported as watertight by \texttt{trimesh}, we first query the analytic volume and area returned by the library. When both values are finite and strictly positive we directly use them as $V$ and $S$.

For non-watertight or numerically unstable meshes we instead use a voxel-based estimate. We voxelize the normalized mesh with a voxel pitch of $0.02$ in all our experiments and obtain a binary occupancy grid. We then regularize this grid using simple volumetric morphology to approximate a single watertight solid and suppress small artifacts. We fill internal cavities by three-dimensional hole filling and apply morphological closing with a spherical structuring element so that thin gaps or cracks in the surface are bridged. We then keep only the single largest six-connected component of occupied voxels and discard small floating pieces, which would otherwise add unstable contributions to the volume and surface area. The volume is computed by counting occupied voxels and multiplying this count by the voxel volume, and the surface area is computed by extracting an isosurface with marching cubes and measuring the triangle area of the resulting mesh. When marching cubes fails we instead count all voxel faces that lie on the boundary between occupied and empty cells.

Finally we compute Compactness from $V$ and $S$ using the definition in Eq.~\ref{eq:compactness} and clamp the result to the range $[0,1]$. If either $V$ or $S$ is non-positive we set $C = 0$.

\paragraph{Normal Anisotropy.}

Normal Anisotropy is computed from the distribution of per-face normals, weighted by face area. For each mesh we take the face normals and the corresponding face areas, normalize the normals to unit length, and assign each face an area weight obtained by dividing its area by the total surface area. We then map every normal to spherical coordinates on the unit sphere using its azimuth angle and its vertical component. This choice makes a uniform grid in this two-dimensional space correspond to patches of roughly equal area on the sphere.

In this two-dimensional space we build a histogram with $K$ bins in total, with $K = 128$ in all our experiments. We choose the number of azimuth and elevation bins so that their product equals $K$, and accumulate the area weights of faces whose normals fall into each bin. The resulting bin values are renormalized to sum to one and define a discrete probability distribution over normal directions. We compute the Shannon entropy of this distribution and normalize it by the entropy of a uniform distribution with $K$ bins. Normal Anisotropy is defined as one minus this normalized entropy as in Eq.~\ref{eq:na}, which by construction lies in the range $[0,1]$.

\paragraph{ModelNet40 category-wise statistics.}
We evaluate Compactness and Normal Anisotropy for all categories in the ModelNet40~\cite{wu20153d} dataset and report the per-category statistics in Fig.~\ref{fig:metric}. Meshes for which loading or metric computation fails are excluded from the averages.

\subsection{User study interface}
We conducted the user study using a web-based interface (Fig.~\ref{fig:userstudy_interface}). For each trial, participants were shown the 2D input image together with an auto-rotating GIF (360-degree render) of one of the 3D models, and were asked to answer the three questions described in the main paper using a 5-point Likert scale. Across 51 participants and 6 methods, this resulted in a total of 918 individual ratings.

\subsection{Dataset}
We collected 2,232 flat images with limited 3D cues and show a few representative examples in Fig.~\ref{fig:data_explain}.

\section{Image-Conditioned 3D Editing}
\paragraph{Editing comparison.}
\begin{figure*}[t!]
  \centering
  \includegraphics[width=1.0\textwidth]{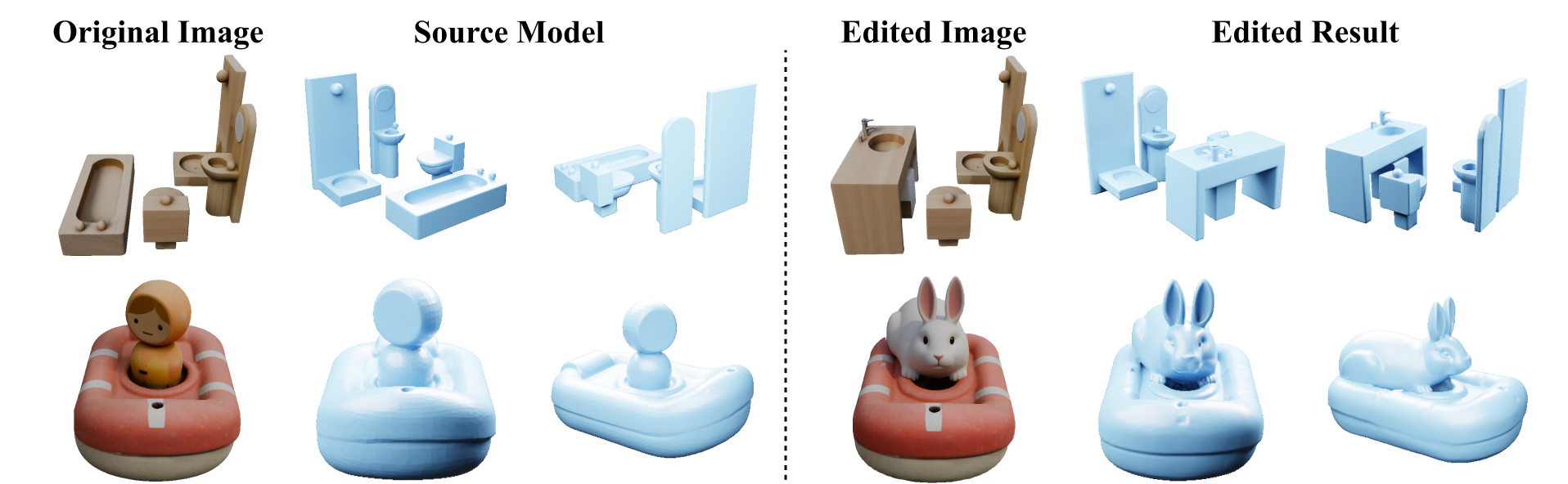}
  \caption{Image-conditioned 3D editing results on the Edit3DBench dataset~\cite{li2025voxhammer}.}
  \label{fig:editing}
\end{figure*}

\begin{figure*}[t!]
  \centering
  \includegraphics[width=1.0\textwidth]{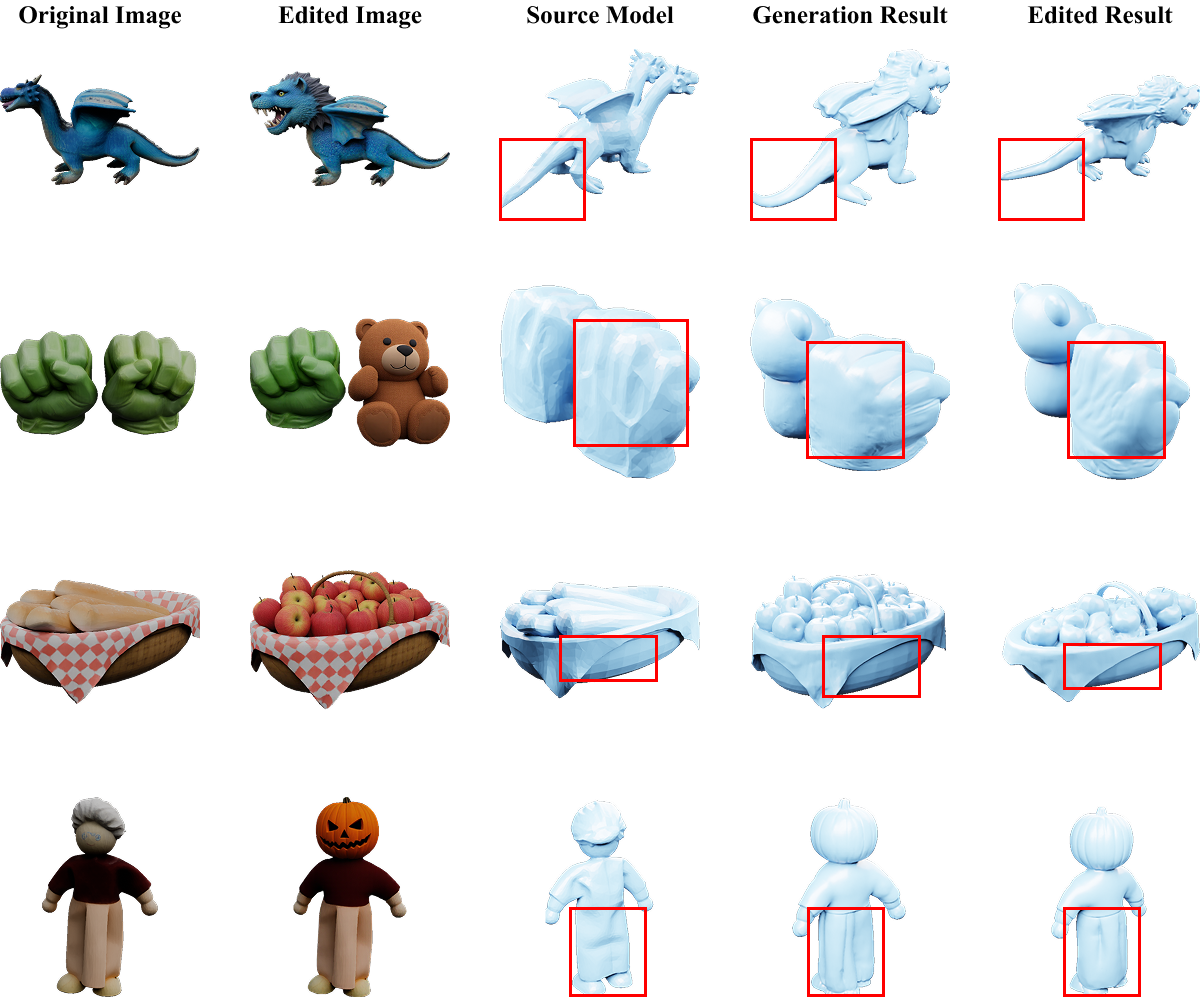}
  \caption{Image-conditioned 3D editing result on the Edit3DBench dataset~\cite{li2025voxhammer}. Generation result denotes the single image-to-3D reconstruction, and the red boxes highlight regions used to check how well the original source geometry is preserved.}
  \label{fig:edit_compare}
\end{figure*}

Our pipeline's latent refinement process can be naturally repurposed for 3D editing. Instead of using the Inflated Prior, we encode a source model to obtain the initial latent $z_0$ and provide an edited image ($y_{\text{edited}}$) as the condition. We then add Gaussian noise according to Eq.~\ref{eq:noising} at the initial noise level $t_0$, reducing fidelity to the source while preserving its coarse structure. Subsequently, denoising under the new edited condition $\psi(y_{\text{edited}})$ steers the latent toward the desired edits. This approach enables high-quality, image-conditioned 3D editing without requiring an explicit 3D editing mask. We demonstrate this application using the Edit3DBench dataset~\cite{li2025voxhammer}, with qualitative results shown in Fig.~\ref{fig:editing}.

In addition, we show qualitative comparisons in Fig.~\ref{fig:edit_compare}. For these experiments, we use Hunyuan3D-2.1~\cite{hunyuan3d2025hunyuan3d} as the 3D latent backbone and compare our editing results against single image-to-3D generations obtained from the same backbone. When editing from a source model, the original geometry is retained in the initial latent representation, so our method preserves the structural details of the source mesh much better than direct generation. For example, the curvature of a tail, wrinkles on the hands, the height of a basket, and wrinkles in clothing remain clearly visible after editing. These results indicate that our approach can modify a 3D mesh to match an edited image while faithfully preserving the source structure.

\section{More Results}

\begin{figure*}[t!]
  \centering
  \includegraphics[width=1.0\textwidth]{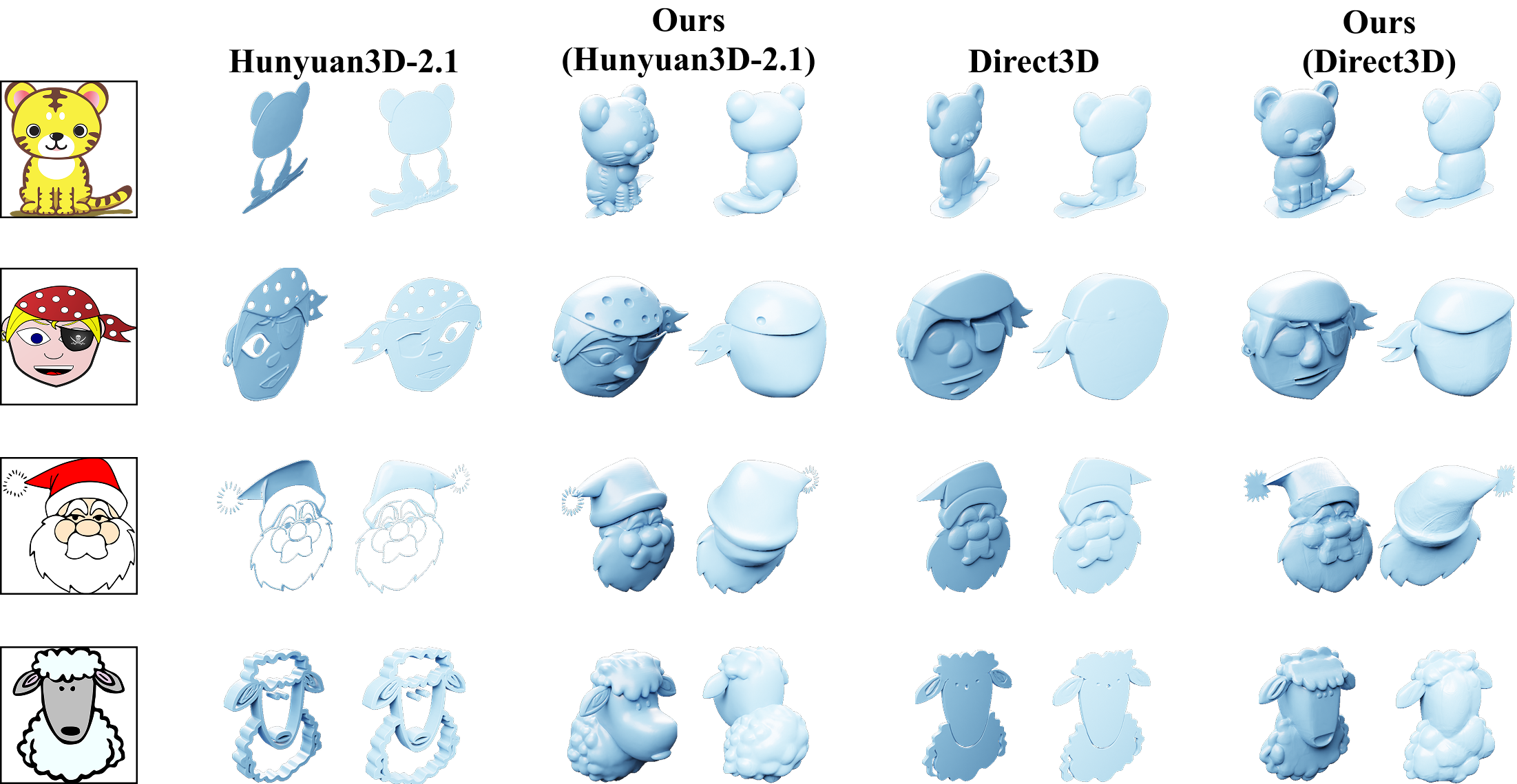}
  \caption{Effect of different 3D latent backbones.}
  \label{fig:backbone}
\end{figure*}

\begin{figure*}[t!]
  \centering
  \includegraphics[width=1.0\textwidth]{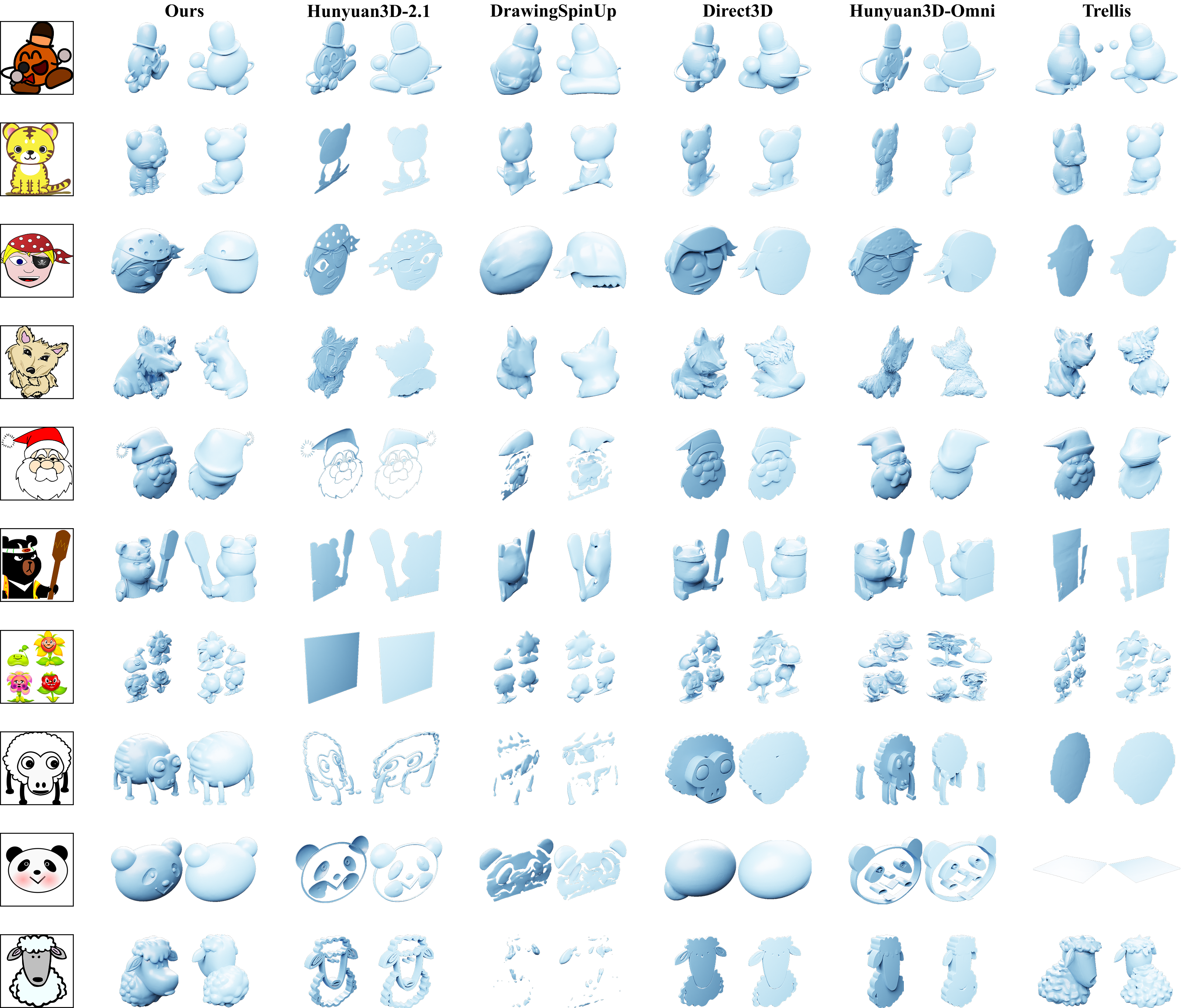}
  \caption{Additional qualitative comparisons of 3D meshes generated from flat images by our method and prior works.}
  \label{fig:more}
\end{figure*}

\begin{figure*}[t!]
  \centering
  \includegraphics[width=1.0\textwidth]{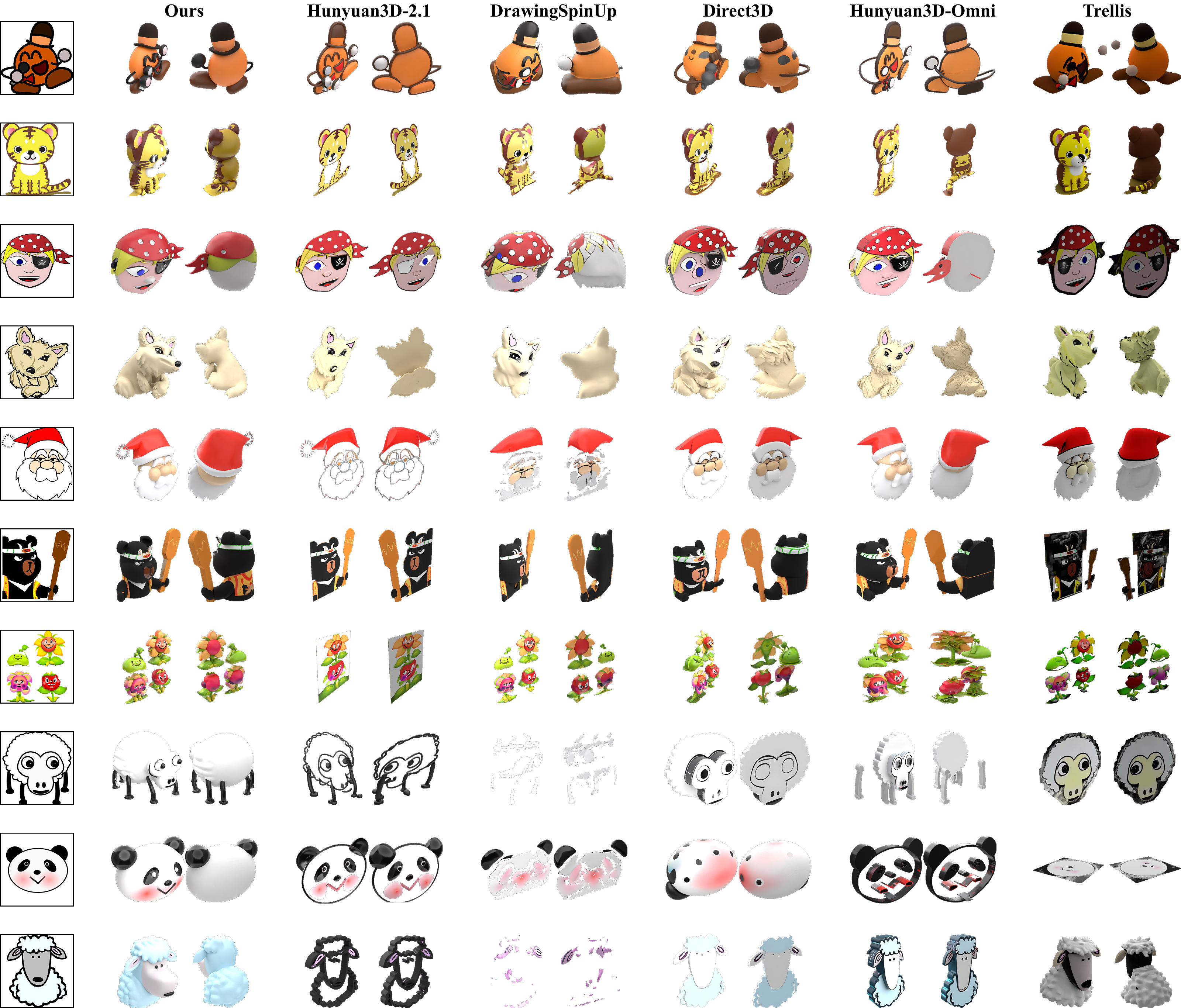}
  \caption{Texture comparison of 3D generations from flat images.}
  \label{fig:more_texture}
\end{figure*}

\subsection{Effect of the backbone.}
The behavior of Stage~2 (3D Latent Refinement) depends on the choice of the pretrained conditional 3D latent generative backbone (Fig.~\ref{fig:backbone}). When Hunyuan3D-2.1~\cite{hunyuan3d2025hunyuan3d} is used as the backbone, the model has been trained to produce rich surface appearance and fine-scale details, so our refinement yields volumetric meshes that preserve these detailed surface structures. In contrast, Direct3D~\cite{wu2024direct3d} is more specialized for recovering the overall shape rather than fine-grained surface structure, and under the same framework it tends to emphasize the global geometry while providing relatively weaker local details. These comparisons indicate that our method consistently produces more volumetric 3D meshes across different backbones, while effectively leveraging the pretrained 3D knowledge encoded in each backbone.

\subsection{More Comparisons}
We conduct an additional experiment that evaluates 3D reconstruction from flat
images, and show representative qualitative results in Fig.~\ref{fig:more}. For all baseline methods, we follow their official configurations as closely as possible to ensure a fair comparison.

For DrawingSpinUp~\cite{zhou2024drawingspinup}, we observed that the default setting sometimes applies overly aggressive thinning on flat inputs, which can prevent any 3D mesh from being generated. In such failure cases, we rerun the method with the thinning step disabled and use these results instead.

Hunyuan3D-Omni~\cite{hunyuan3d2025hunyuan3domni} requires a 3D bounding box as a conditioning input. To obtain the best possible performance, we set the $x$ and $y$ extents from the normalized foreground bounding box of the input image and fix the depth extent $z$ to 0.6 in all our experiments.

\paragraph{Texture comparison.}
Trellis~\cite{xiang2024structured} directly generates textured 3D meshes, whereas all other methods produce geometry-only meshes. To compare textured appearance, we apply the Hunyuan3D-Paint pipeline from Hunyuan3D-2.1~\cite{hunyuan3d2025hunyuan3d} to each geometry-only mesh, using the original 2D image as the conditioning input for texture synthesis. The resulting textured meshes are shown in Fig.~\ref{fig:more_texture}.


\end{document}